\documentclass[lettersize,journal]{IEEEtran}
\usepackage{amsmath,amsfonts}
\usepackage{algorithmic}
\usepackage{algorithm}
\usepackage{array}
\usepackage{balance}
\usepackage[caption=false,font=normalsize,labelfont=sf,textfont=sf]{subfig}
\usepackage{textcomp}
\usepackage{stfloats}
\usepackage{url}
\usepackage{verbatim}
\usepackage{graphicx}
\usepackage{cite}
\usepackage{hyperref}
\usepackage{amsmath,amssymb,amsfonts}
\usepackage{algorithmic}
\usepackage{graphicx}
\usepackage{textcomp}
\usepackage{multirow}

\usepackage{xcolor}
\usepackage[bottom]{footmisc} 
\usepackage{flushend} 

\usepackage{wrapfig}
\usepackage{xcolor,colortbl}
\definecolor{Gray}{gray}{0.25}
\definecolor{red}{rgb}{1.00,0.00,0.00}
\definecolor{blue}{rgb}{0.00,0.00,1.00}
\definecolor{green}{rgb}{0.2,0.80,0.0}
\definecolor{yellow}{rgb}{0.5,0.5,0.0}

\newcommand{\cblue}[1] {\textcolor{blue}{#1}}

\usepackage{graphicx}
\usepackage{amsmath,amssymb,amsfonts}
\usepackage{hyperref}

\usepackage{multirow}
\usepackage{mathtools}
\usepackage{arydshln}
\usepackage{xcolor}
\usepackage{wrapfig}
\usepackage[bottom]{footmisc} 

\begin{document}

\title{Simultaneous Multi-View Object Recognition \\ and Grasping in Open-Ended Domains}
\author{\IEEEauthorblockN{\textsuperscript{} Hamidreza Kasaei$^*$, Mohammadreza Kasaei, Georgios Tziafas, Sha Luo, Remo Sasso}
\thanks{
All authors are with Department of Artificial Intelligence, Bernoulli Institute, University of Groningen, 9747 AG, The Netherlands. \newline $^*$Corresponding author: hamidreza.kasaei@rug.nl}}
\markboth{}%
{Simultaneous Multi-View Object Recognition and Grasping in Open-Ended Domains}
\maketitle

\begin{abstract}

To aid humans in everyday tasks, robots need to know which objects exist in the scene, where they are, and how to grasp and manipulate them in different situations. Therefore, object recognition and grasping are two key functionalities for autonomous robots. Most state-of-the-art approaches treat object recognition and grasping as two separate problems, even though both use visual input. Furthermore, the knowledge of the robot is fixed after the training phase. In such cases, if the robot encounters new object categories, it must be retrained to incorporate new information without catastrophic forgetting. In order to resolve this problem, we propose a deep learning architecture with an augmented memory capacity to handle open-ended object recognition and grasping simultaneously. In particular, our approach takes multi-views of an object as input and jointly estimates pixel-wise grasp configuration as well as a deep scale- and rotation-invariant representation as output. The obtained representation is then used for open-ended object recognition through a meta-active learning technique. We demonstrate the ability of our approach to grasp never-seen-before objects and to rapidly learn new object categories using very few examples on-site in both simulation and real-world settings. A video of these experiments is available online at:  \href{https://youtu.be/n9SMpuEkOgk}{\cblue{\texttt{https://youtu.be/n9SMpuEkOgk}}} 
\end{abstract}


\section{Introduction}

\begin{figure}[!t]
    \centering
    \includegraphics[width=\linewidth]{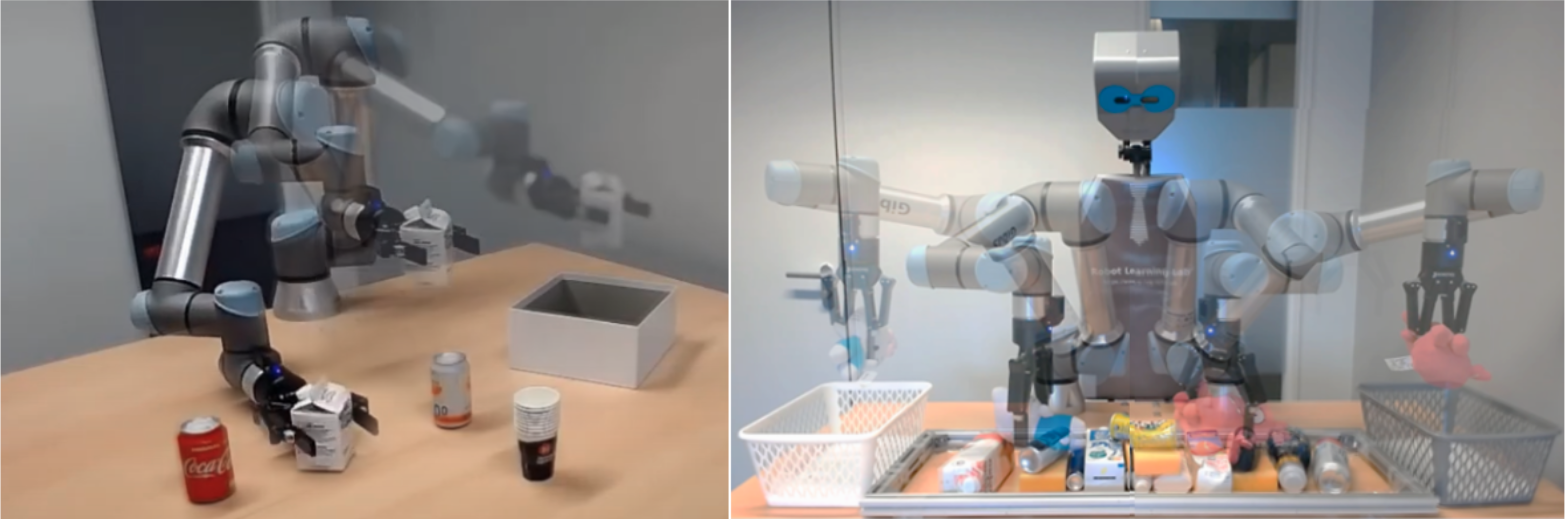}
    \vspace{-5mm}
  \caption{\small To accomplish various tasks successfully, (\textit{left}) a robot must understand which objects exist in the scene, where they are, and where to move its gripper to pick up the target object. Our approach allows robots to learn new object categories using very few instances on-site. (\textit{right}) The proposed approach also allows the robot to predict stable grasp configurations for a diverse set of objects in highly cluttered scenarios. }
  \label{clear_table_task}
\vspace{-1mm}
\end{figure}

  \IEEEPARstart{T}{he} necessity of using robots in human-centric environments has led to fast progress in the field of machine learning, computer vision, and robotics~\cite{wang2021spatial}\cite{Yu_2020_CVPR}\cite{fang2020graspnet}. To assist humans in various daily tasks ( e.g., clear table), a robot needs to know which kinds of objects exist in a scene, where they are, and how to grasp and manipulate the target object. Robots operating in such dynamic environments frequently faces isolated never-seen-before objects or a pile of objects (see Fig.~\ref{clear_table_task}). Therefore, they should be able to learn new object categories on-site from very few training examples while retaining their previous knowledge. Recent breakthroughs on object perception and manipulation often use deep learning techniques. While deep learning is a very powerful tool, there are several limitations to using deep neural network in open-ended domains. First, deep learning approaches are data-hungry approaches as learning a new skill/concept usually requires hundreds to thousands of sufficiently similar training instances. Therefore, the training process is computationally expensive and slow. Second, the model is trained once all data has been gathered and its performance strongly dependents on the quality and quantity of training data. Often, the learned models  do not generalize well to never-seen-before objects, and training with limited data leads to poor performance. Deep learning approaches are also prone to catastrophic forgetting~\cite{kirkpatrick2017overcoming}. 

\begin{figure*}[!t]
    \vspace{-5mm}
    \centering
    \includegraphics[width=\linewidth]{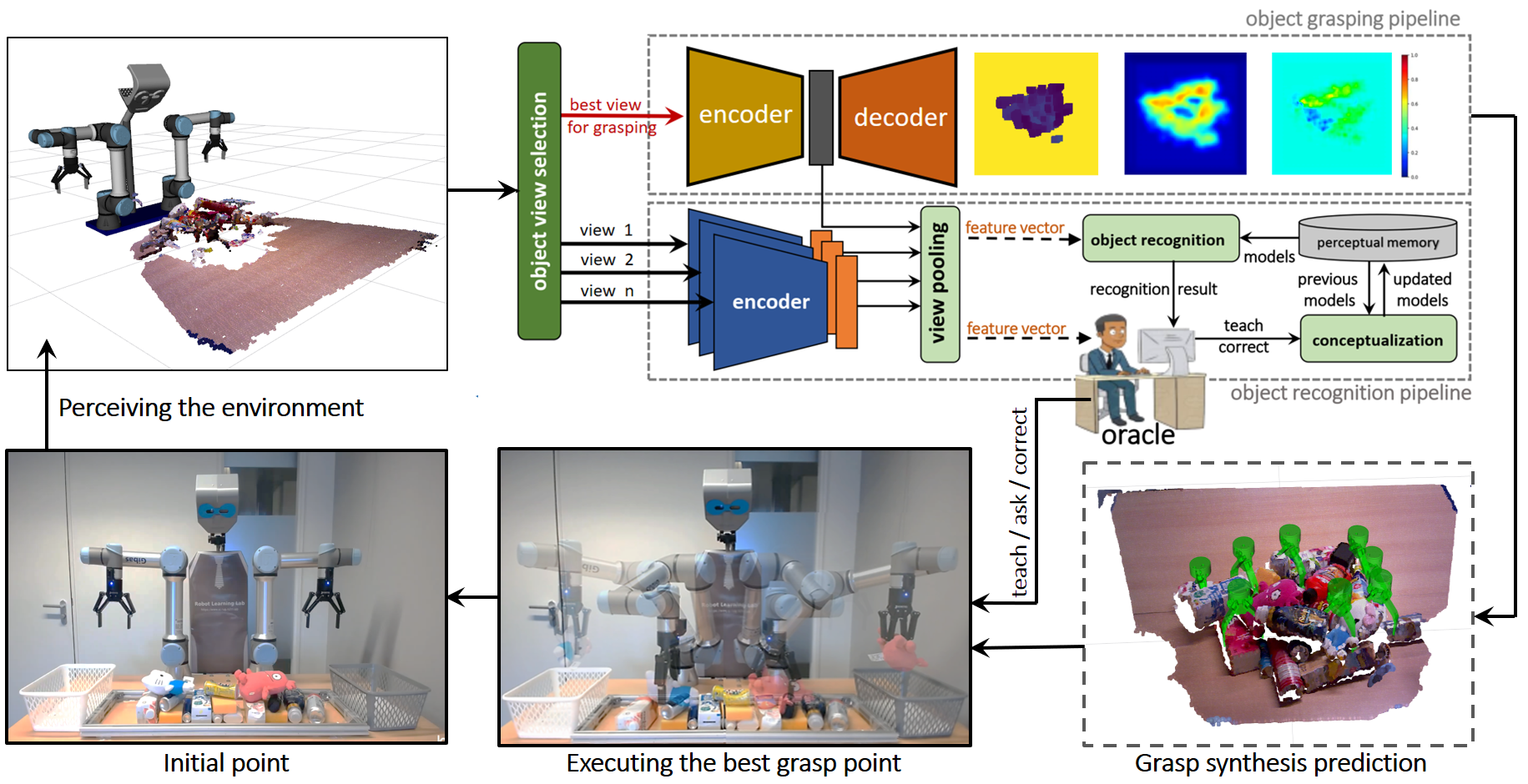}
    \vspace{-4mm}
  \caption{\small  We propose a deep learning approach with an
augmented memory capacity to handle multi-view object grasping and recognition tasks simultaneously.  First, multiple RGB-D views of a given object are generated from different perspective. All RGB views contributed equally to open-ended object recognition, while the depth view with maximum entropy is used for grasping and encoding the geometrical feature of the object.  The depth view of the object is then fed to the grasp network to obtain a pixel-wise grasp configuration and a compact representation. All RGB views of the object are passed into the Vision Transformer and fused together using a pooling function to form a feature vector from RGB views of the object. The depth representation and the RGB representation are then concatenated to form a global representation of the object. The obtained representation is finally used for the downstream open-ended learning task. 
  }
  \label{system_overview}
\end{figure*}

  In this paper, we aim to address these limitations by making robots capable of learning the category label of objects in an open-ended manner through interaction with non-expert users. In particular, the robot has the ability to ask users to label some of the training instances in which it is unsure about. This way, the robot is able to update its knowledge incrementally rather than having to retrain from scratch when a new instance is introduced or a new category is taught. Furthermore, apart from robot self-learning, non-expert users could interactively guide the robot by teaching new concepts, or by correcting insufficient or erroneous concepts. We propose to study this problem at the crossroad of deep learning and meta-active learning. An overview of the proposed approach is shown in Figure~\ref{system_overview}. We develop an external-memory equipped deep learning approach capable of producing grasp configuration and a compact object representation for a given object. The obtained representation is scale- and rotation-invariant, informative, and stable, and designed with the objective of supporting accurate 3D object recognition in open-ended domains. More specifically, our approach combines the best of two worlds: the ability to slowly learn an object-agnostic grasping and a compact object representation function, via gradient descent, and the ability to rapidly learn about new categories using very few examples, via meta-active learning. Furthermore, addressing object recondition and grasping is important for real-time robotic applications, especially if there are resource constraints. Our contributions are three-fold:
\begin{itemize}

\item  We develop a deep learning architecture with an augmented memory capacity to handle object grasping and continual object recognition  simultaneously. 
    \item We develop a probabilistic learning method to handle 3D object recognition in open-ended domains;
    \item To assess the effectiveness of proposed approach, we perform extensive sets of experiments in both simulation and real-robot settings. Our method enables a robot to learn about new object category using, on average, less than five instances per category and achieve $95\%$ object recognition accuracy and above $91\%$ grasp success rate on (highly) cluttered scenarios in both simulation and real-robot experiments.
\end{itemize}

\section{Related work}

Although an in-depth review is beyond the scope of this work, we discuss recent efforts in three main categories: object grasping, object recognition, and active learning.

\textbf{Object grasping} -- Earlier methods on object grasping were mainly based on hand-crafted features~\cite{bohg2013data}. In recent studies much attention has been given to Convolutional Neural Networks (CNN)~\cite{lenz2015deep}\cite{mahler2017dex}\cite{morrison2018closing}\cite{klokov2017escape}\cite{kanezaki2018rotationnet}\cite{kumra2020antipodal}\cite{breyer2020volumetric}\cite{mousavian20196}. 
In particular, CNNs have been applied successfully for empirical object grasping methods. In such approaches, the grasps are classified and ranked using a CNN, after which a robot executes the highest-ranked grasp such as in \cite{mahler2017dex}. One of the biggest bottlenecks with recent deep learning-based object grasping approaches is the execution time. Some of the deep-learning-based approaches take a very long time to sample and rank grasp candidates (e.g.,~\cite{lenz2015deep}\cite{mahler2017dex}), while others first need to explore the environment to acquire a full model of the scene and then generate point-wise 6D grasp configuration (e.g., Volumetric Grasping Network (VGN)~\cite{kumra2020antipodal}. These approaches mainly use in open-loop control scenarios and are not suitable for closed-loop scenarios. Morrison et al.~\cite{morrison2018closing} proposed the Generative Grasping CNN (GG-CNN), a small neural network, which generates pixel-wise grasp configurations for a given single-modal image (depth-only). Kumra et. al.,~\cite{kumra2020antipodal} developed GR-ConvNet, a \textit{large} deep network that generates pixel-wise grasp configurations using multi-modal data. Contrary to our approach, VGN~\cite{breyer2020volumetric} and GG-CNN~\cite{morrison2018closing}, which only use depth data, GR-ConvNet combines color and depth information. 

Similar to our approach, GG-CNN is designed to be used for real-time closed-loop control using visual feedback. Unlike GG-CNN, our approach works in an eye-to-hand system, where the robot considers an entire scene and not just a narrow top-down view. Our approach generates a grasp map per \textit{object} while GG-CNN, GR-ConvNet, and DexNet generate a grasp map per \textit{scene}.  Unlike our approach and VGN, GR-ConvNet and GG-CNN both work in top-down camera settings and mainly focused on solving 4DoF $(x, y, z, \phi)$ grasping, where the gripper is forced to approach objects from above. A major drawback of these approaches is inevitably restricted ways to interact with objects. 

\textbf{Object recognition} -- Nowadays visual recognition systems are often designed based on CNN, where the number of classes is known in advance as prior information~\cite{bochkovskiy2020yolov4}. Although these approaches work well in static closed set environments, they easily fail when facing an out-of-distribution instance (e.g., fooling image) by predicting a ``\textit{known}'' label with high confidence~\cite{bendale2016towards,subramanya2019fooling}. Some researchers tried to handle this limitation by incorporating an ``\textit{unknown}'' class~\cite{da2014learning}. Although these approaches can detect ``\textit{unknown}'' objects to some extent, they cannot learn about new categories due to catastrophic forgetting (learning about new object categories leads to forget previously learned categories)~\cite{kirkpatrick2017overcoming, scheirer2014probability}. In general, deep learning approaches for 3D object recognition can be categorized into three different categories depending on their input. First, there are volume-based approaches \cite{wu20153d,maturana2015voxnet}, where the object is represented as a 3D voxel grid and then fed to a CNN with 3D filter banks. Second, there are pointset-based approaches~\cite{klokov2017escape}, which work directly on the 3D point clouds. The final category is view-based approaches, which are used in this research. These approaches appear to be most effective in 3D object recognition, as shown by \cite{kanezaki2018rotationnet}, \cite{qi2016volumetric}\cite{shi2015deeppano}. In such approaches, 2D images are extracted from the 3D representation by projecting the object’s points onto 2D planes \cite{su2015multi,shi2015deeppano}. H. Su et al., \cite{su2015multi} developed a system that learns to recognize 3D shapes from a collection of their rendered views on 2D images, for which multiple view-wise CNN features were used. Another approach, by \cite{kanezaki2018rotationnet}, takes multi-view images of an object as input and jointly estimates its pose and object category label using a CNN. Our research relates to these works as both use multi-view representations of 3D objects to learn deep features. However, we trained an autoencoder to generate a grasp map as well as a compact deep representation for a given object. The learned deep features are used for open-ended object category learning and recognition. Unlike these approaches, the set of object categories to be learned is not completely known in advance in our approach, and the model does not know which additional objects it will have to learn, which observations will be available, and when they will be available to support the learning.

\textbf{Active learning (AL)} -- In recent years, AL methods have been gaining much attention to overcome the aforementioned limitations~\cite{sener2017active, Aggarwal_2020_WACV, Siddiqui_2020_CVPR,gal2017deep}, but few AL methods target the problem of open-ended learning~\cite{OrthographicNet, kasaei2016hierarchical}. In particular, most AL approaches, first sample a subset of training examples from a pool of unlabeled data using an acquisition function based on either '\textit{uncertainty}' measures (entropy, variance, and etc.) or density/geometric similarly measures in feature space (i.e., sampling diverse instances by considering the similarities among training data). An oracle is then asked to label the selected samples. Finally, the model is incrementally trained or re-trained from scratch to incorporate new information without catastrophic interference. These approaches are incremental by nature but not open-ended since the number of categories is pre-defined and the main objective is to update the model of known categories by finding minimally required training examples to reach a certain classification accuracy. Moreover, unlike these approaches, we formulate the AL to learn from online robot's observation and not from a set of training data. More specifically, instead of selecting a set of instances that represents the entire training dataset, we want to select a set of training samples that best represents the novel classes. We also update the model of known categories only when it is necessary. This way, we mainly use our limited labeling budget to learn about new object categories and update the model of known classes when necessary.

\section{Object Representation and Grasp Learning}

We formulate object representation and grasp synthesis as a learning problem. In particular, we intend to learn a function that receives a collection of rendered images of a 3D object as input, and returns (\textit{i}) a compact, scale- and rotation-invariant representation, (\textit{ii}) the best direction for approaching the target object, and (\textit{iii}) a grasp map representing per-pixel grasp configuration for a selected view. 
\subsection {Generating Multi Views of 3D Objects}

\begin{figure}[!t]
  \centering
  \includegraphics[width=0.9\linewidth, trim = 0cm 0cm 0cm 0cm clip=true]{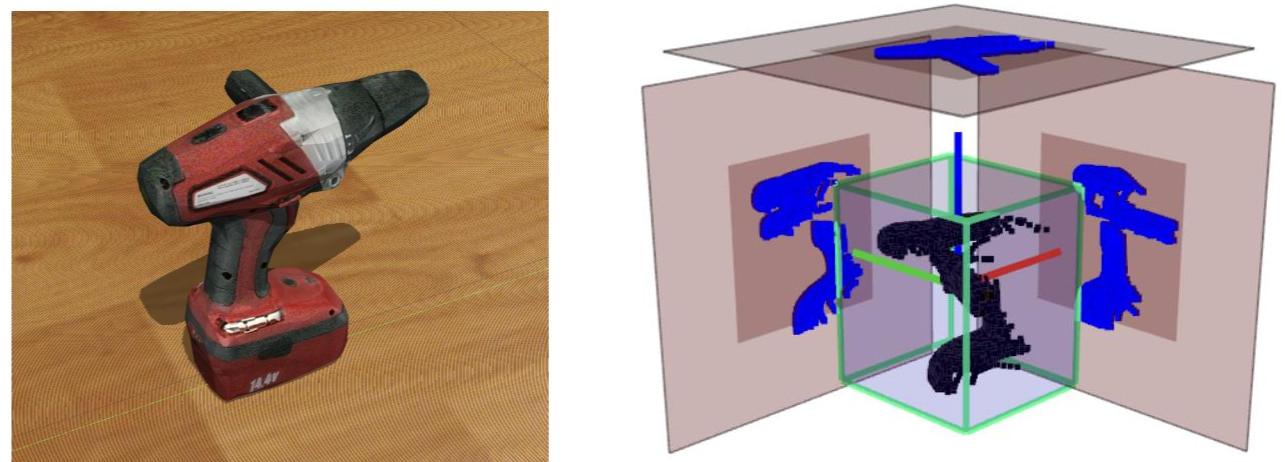}
  \caption{(\textit{left}) a cordless drill; (\textit{right}) the partial point cloud of the object, its local reference frame, bounding box, and three projected views of the drill. In each projection, the darker area shows the image size for the object representation task, and the lighter area represents the size of the image for the object grasping task.}
  \label{projections}
\end{figure}

A point cloud consists of a set of points, $p_i : i \in \{1,\dots,n\}$, where each point is described by its 3D coordinates $[x, y, z]$. To render 2D depth images from a 3D object, we set \textit{``virtual''} cameras around the target object, whose Z axes point towards the centroid of the object. Towards this goal, we first compute the geometric center of the object, which is defined as the arithmetic mean position of all its points. Afterwards, we construct a local reference frame for the object by performing eigenvalue decomposition analysis on the normalized covariance matrix, $\Sigma$, of the object, i.e., $\Sigma\textbf{V}=\textbf{EV}$, where $\textbf{E} = diag(e_1, e_2, e_3)$ contains the descending sorted eigenvalues, and $\textbf{V} = (\vec{v}_1, \vec{v}_2, \vec{v}_3)$ shows the eigenvectors. We consider the first two largest eigenvectors, $\vec{v}_1$ and $\vec{v}_2$, as $\textbf{X}$ and $\textbf{Y}$ axes respectively, and define the $\textbf{Z}$ axis as the cross product of $\vec{v}_1 \times \vec{v}_2$. The object is then transformed to be placed in the reference frame (see Fig.~\ref{projections} \textit{right}).

From each virtual camera pose, we map the point cloud of the object into a depth image using the z-buffering and orthogonal projection methods~\cite{liu2019soft} regardless of how accurate/complete the point cloud of the object is. In particular, we first project the object to a square plane centered on the camera’s center. 
Note that the size of the projection square area, $l \times l$, is an important factor for both object representation and object grasping tasks. In the case of object representation, we define the size of projection relative to the size of the object for producing a scale-invariant object representation. In particular, the size of the projection plane is defined as $l_p \times l_p$ dimension, where $l_p$ is the largest side of axis-aligned bounding box of the object. Since the grasp configurations depend on the pose and size of the target object, a view of the object should not be scale-invariant. Therefore, we consider a fixed size projection plane for grasping ($l_g \times l_g$).
In our setup, $l_g$ parameter is set to $0.45$m. The projection area is then divided into $k \times k$ square bins, where each bin is considered a pixel. An illustrative example of this procedure is provided in Fig.~\ref{projections}. 

\subsection{Virtual Viewpoint Setups}

The number of views for each object is an important parameter for both object grasping and object recognition. 
Although viewpoint setup can be any arbitrary choice, we consider three setups in this work: (\textit{left}) orthographic projections, i.e., $\{v_i\}_{i=1} ^ 3$, (\textit{center}) an orbit elevated by $\phi$ (similar to MVCNN~\cite{su2015multi}), and (\textit{right}) a sphere viewpoints setup, which is similar to the previous setup but with multiple elevation levels. {The setup of orthographic projection has been explained in the previous subsection.} For the orbit viewpoint setup, we place virtual cameras around the \textbf{Z} axis at intervals of $\alpha$, elevated by a fixed $\phi$. Therefore, the number of views for a given object is set to $\{v_i\}_{i=1}^{V = \frac{360}{\alpha}}$. In the case of sphere viewpoint setup, instead of having a fix elevation, we placed virtual cameras at multiple elevation levels, $\beta$, with the interval of [$-90^{\circ}$, $90^{\circ}$]. Therefore, we capture $V = \frac{360}{\alpha} \times \frac{180}{\beta}$ views for a given object. We have optimized $\phi$, $\alpha$, and $\beta$ parameters to obtain a good balance between object recognition accuracy and computation time {(see section~\ref{exprimental_result}-A)}. It should be noted that object recognition treats all views equally important, while object grasping ranks views based on visibility, reachability, and collision-free metrics.

\subsection{View Selection for Grasping}
View selection is crucial to make a multi-view approach computationally efficient. Although it is possible to pass all the views of the object into the network and then execute the grasp with a maximum score that is kinematically feasible (Fig.~\ref{different_grasps} \textit{left}), such approaches are computationally expensive. In contrast, choosing a view that covers more of the target object's surface will not only reduce the computation time but also increase the likelihood of grasping the object successfully. Information theory provides a range of metrics (variance, entropy, etc.) from which the expected information gain can be calculated. Among these metrics, viewpoint entropy is a good proxy for expected information gain~\cite{thrun2002probabilistic}. In particular, viewpoints that observe the area of high entropy are likely to be more informative than those that observe low entropy areas. Therefore, we formulate our view ranking procedure using viewpoint entropy, which considers both the number of occupied pixels and the pixels' values. In particular, we calculate the entropy of a normalized projection view, $v$, by $H (v) = - \sum_{k=1}^{k^2} p_k \log_2(p_k)$, where $p_k$ is the normalized value of pixel $k$, and $\sum_k p_k = 1$. The view with highest entropy is considered as the best view for grasping and then fed to the network to predict pixel-wise grasp configuration and encode the geometrical feature of the object. The gripper approaches the object from an orthogonal direction to the projection. 

\begin{figure}[!t]
  \begin{tabular}{c|c|c}
    \hspace{-7mm}
    \includegraphics[width=0.3\linewidth,trim = 0cm -0.3cm 0cm 0cm clip=true]{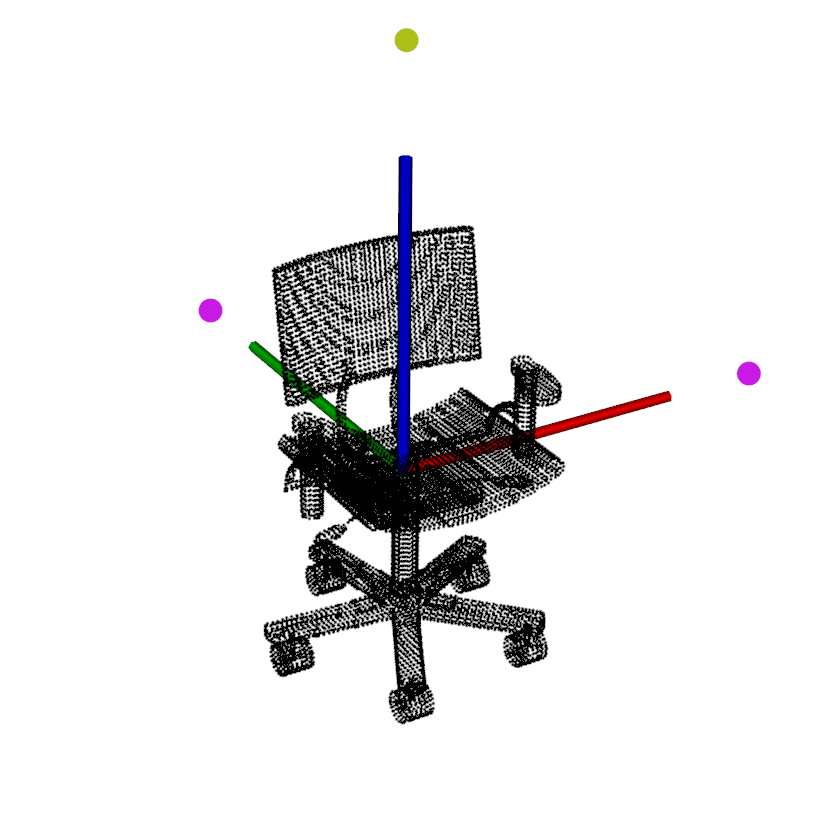}
  &    \includegraphics[width=0.3\linewidth, trim = 0cm -1.7cm 0cm 0cm clip=true]{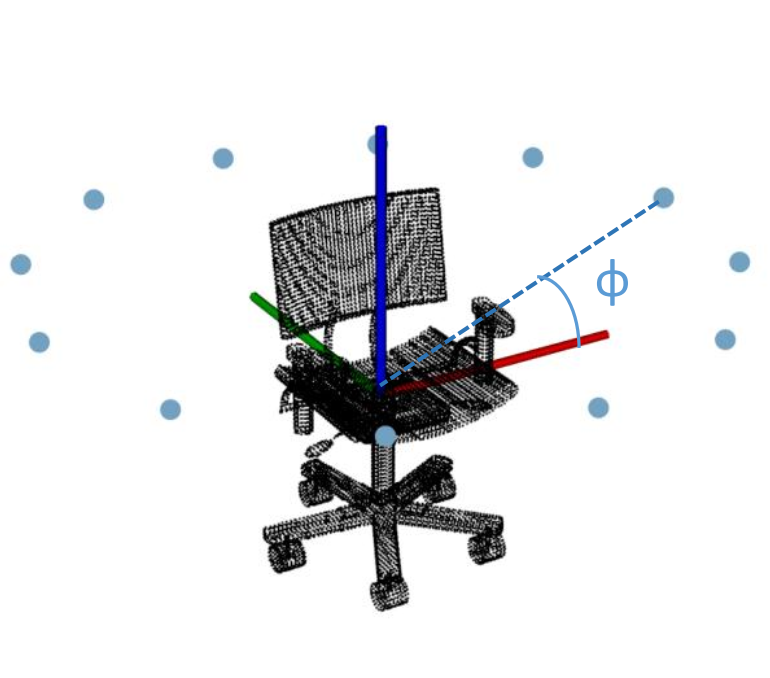}
  &    \includegraphics[width=0.3\linewidth, trim = 0cm 0cm 0cm 0cm clip=true]{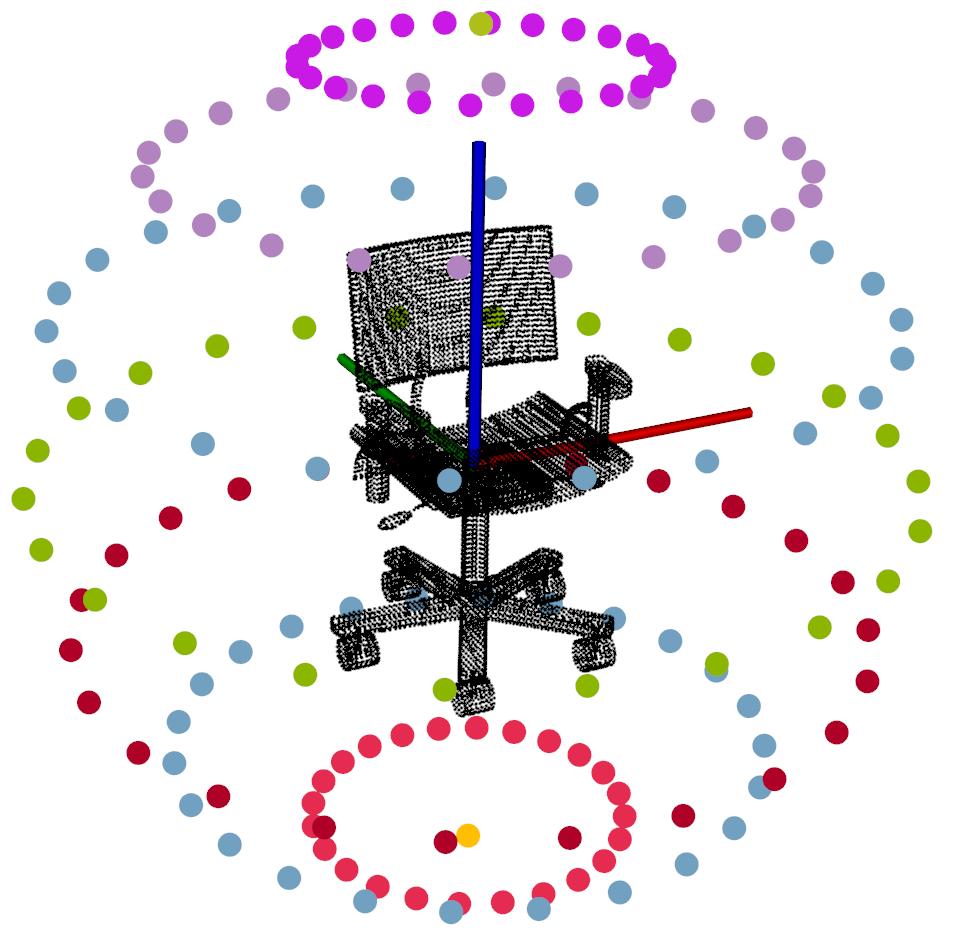}\\
      \scriptsize \textit{orthographic} & \scriptsize \textit{orbit elevated by} $\phi$ & \scriptsize \textit{sphere}
  \end{tabular}
  \caption{\small Illustration of three viewpoint setups used in this study. In all cases, distances between cameras and the center of the target object are constant and elevation levels are shown by colors. }
  \label{viewpoint_setups}
\end{figure}

\subsection {Network Architecture}
\label{network}



We intend to develop a small grasp network that maps an input depth image to multiple outputs, including a compact deep representation for encoding the geometrical feature of the object for recognition purposes, and a set of images representing pixel-wise antipodal grasp configurations, as we are interested to use the network in real-time robotic applications with limited resources.
To encode the textural feature of the object, we use a Vision Transformer (ViT)~\cite{dosovitskiy2020image}, pre-trained on ImageNet-1k~\cite{deng2009imagenet}. The RGB view of the object are independently fed into the ViT, and the resulting features are fused into a single vector representation using a pooling function. The depth and the RGB representations are then concatenated to form a single global representation for the given object. The obtained representation is then used for downstream open-ended object category learning and recognition task.

\begin{figure}[!t]
    \centering
    \includegraphics[width=\linewidth]{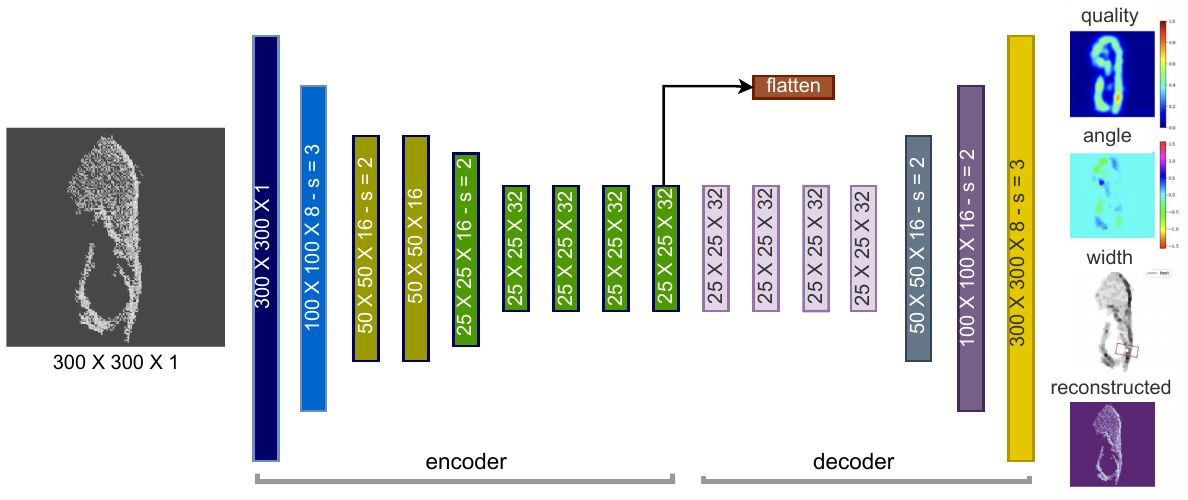}
    \vspace{-7mm}
    \caption{The overall architecture of the proposed grasp network. The network received a depth image as an input and produce a reconstructed view and a pixel-wise grasp configuration as output. The size of image and number of filters after applying convolution is represented as I $\times$ I $\times$ F. Moreover, the strides is represented by s, we do not indicate the s = 1. The last layer of encoder is flattened and considered as deep representation of the input image.}
    \label{network_architecture}
\end{figure}

The grasp network receives a depth image with height $H$ and width $W$ as input, $\mathcal{X} \in \mathbb{R}^{H \times W}$, and returns multiple outputs including: (\textit{i}) a reconstructed image $\hat{\mathcal{X}}$, and (\textit{ii}) a pixel-wise grasp configuration map, $\textbf{G}$, which is represented by rotation, width, and quality images $(\boldmath{\phi}, \textbf{W}, \textbf{Q})$ $\in \mathbb{R}^{H \times W}$, i.e., $f_\theta: \mathcal{X} \rightarrow \mathbf{\mathcal{Y}}$, where $\mathcal{Y} = [\textbf{G}, \hat{\mathcal{X}}]$. We have considered the dense autoencoder and image reconstruction loss to force the network to learn a compact deep representation in the bottleneck layer in an unsupervised manner. The obtained representation is used for object recognition purposes (i.e., meta learning, as we learn about new categories using the output of other learning method). The overall architecture of the network is depicted in Fig.~\ref{network_architecture}. The encoder part of the network is composed of an input layer followed by eight convolutional layers, while the decoder part is composed of seven deconvolutional layers. We use Rectified Linear Unit (\texttt{ReLU}) as the activation function of all layers. Except for the last deconvolution layer, we have added a batch normalization layer after each convolution and deconvolution layers to stabilize the learning process and reduce the number of training epochs by keeping the mean and standard deviation of output close to 0 and 1, respectively. We use the same padding in all convolution and deconvolution layers to make the input and output be of the same size. The output of the last convolution layer is flattened and considered as the deep representation of the object~(see Fig.~\ref{system_overview}). The network is trained in an end-to-end manner using the Huber loss (see the details in Sec. \ref{network_analysis}). 

\subsection{Grasp Execution}
\begin{figure}[!b]
    \vspace{-0mm}
    \centering
        \includegraphics[width=\linewidth]{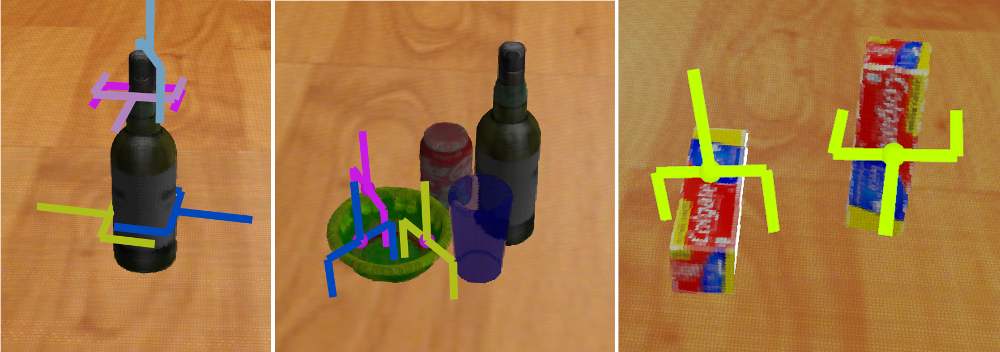}
  \caption{Examples of grasping objects in different situations: (\textit{left}) predicted grasp configurations for orthographic views of the object; (\textit{center}) grasp prediction for a clutter scene that are both kinematically feasible and collision free; (\textit{right}) the best grasp configuration for grasping a Colgate object in two different situations.}
\vspace{-2mm}
\label{different_grasps}
\end{figure}

After obtaining the output of the network for a given input, $f_\theta (I) = \mathbf{G}$, the best grasp configuration, $\operatorname{g^*}$, is defined as the one with maximum quality, and its coordinate shows the center of grasp, i.e., $(u,v) \leftarrow \operatorname{g^*} = \operatorname*{argmax}_\mathbf{Q} ~ \mathbf{G}$. 
Additionally, the distance that the robot needs to travel within the configuration space, as well as the pose of other objects in the scene are considered to verify the feasibility of executing the grasp. Additional constraints due to the kinematic chain of a manipulator are beyond the scope of this work and can be handled by trajectory optimization techniques. When a pile of objects is involved, grasping from above has a clear advantage (e.g., no collisions), but when there are isolated  and cluttered objects, it completely depends on the object's position. Three examples of grasping objects in different situations are shown in  Fig.~\ref{different_grasps}

In this work, we represent the grasp point as a tuple, $g* = \langle (u, v), \phi_i, w_i, q_i\rangle$, where $(u, v)$ represents the center of grasp in virtual image coordinates, $\phi_i$ indicates the rotation of the gripper around the depth axis, $w_i$ represents the necessary width of the gripper, and the success probability of the grasp is represented by $q_i \in [0,1]$. The depth value of the grasp point is determined by the minimum depth value of its surrounding neighbors within a distance of $\Delta$. We set $\Delta = 2.5$cm based on the size of the robot's finger. Afterwards, we transform the coordinates of the grasp point from the virtual view of the object to the reference frame of the object and instruct the robot to perform the grasp action.

\section{Open-ended Object Category Recognition}
\label{object_category_learning}
\begin{figure}[!t]
    \centering
    \includegraphics[width=\linewidth]{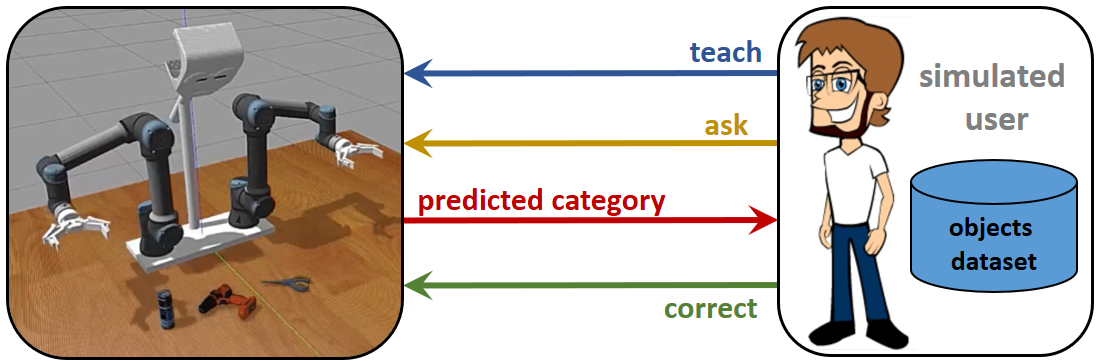}
    \vspace{-7mm}
    \caption{Abstract architecture for interaction between the simulated user and the robot: The simulated user utilizes the "teach" action to teach the robot a new object category; the "ask" action is used to assess the robot's performance on previously learned categories, and the correct action is used to provide corrective feedback when misclassification occurs. }
    \label{simulated_user_arch}
\end{figure}

Most active learning methods do not perform well in open-ended domains since they need to know the number of categories in advance. In open-ended learning scenarios, the number of classes is updated over time, based on the robot's observations, experiences, and interactions with human users. In other words, instead of sampling and labeling the training data in advance, we propose to iteratively and adaptively choose which training instance should be labeled next. In this study, we follow an active learning scenario by identifying the need for teaching a new category or by letting the user provides corrective feedback to learn the model as quickly as possible (see Fig.~\ref{simulated_user_arch}). In particular, we provide three basic actions for the user to either \texttt{teach} the robot about new categories or \texttt{correct} the robot on errors by providing feedback. These actions consist of the following: (\textit{i}) \textbf{ask}: to check the prediction accuracy of an object category model, (\textit{ii}) \textbf{teach}: to introduce a new object category using a set labeled samples, and (\textit{iii}) \textbf{correct}: to improve an object category model using a new instance. The \emph{teach} and \emph{correct} actions lead the robot to initialize a new class or to modify a known class incrementally using a particular instance the current classifier is the least certain about. In particular, we are interested in learning a probabilistic model for each object category, $\textbf{C}$, using very few labelled data, $\mathcal{L}_t = \{\textbf{x}_1, \dots, \textbf{x}_{n_t}\}$, where $n_t$ is the number of seen instances until time $t$, and each instance, $\textbf{x}$, is fed into the encoder network and represented as a d-dimensional feature vector, $[x_1, \dots, x_d]$ where $\sum_{i}^d x_i = 1$. Therefore, we represent an object category as a tuple ~$\textbf{C}_k = \langle~n_k,~\textbf{a}_k,~\operatorname{P}(\textbf{C}_k),~[\operatorname{P}(x_1|\textbf{C}_k),\dots, \operatorname{P}(x_d|\textbf{C}_k)]~\rangle$, where $n_k$ represents the number of seen instances in category $k$ and $\textbf{a}_k$ is a vector of accumulator for category $k$. In particular, $a_{ki}$ is the probability accumulation of $i^{th}$ element of all instances of category $\textbf{C}_k$ and $|\textbf{a}| = |\textbf{x}|$.
$\operatorname{P}(\textbf{C}_k)$ shows the prior probability of category $\textbf{C}_k$ (i.e., $\operatorname{P}(\textbf{C}_k) = n_k/N$, where $N$ is the number of seen instances in all categories).
In this work, we consider the probability of each element of feature vector independently, regardless of any possible correlations with the other elements. This way, the $\operatorname{P}(\textbf{C}_k)\operatorname{P}(\textbf{x}|\textbf{C}_k)$ is equivalent to the joint probability model. Therefore, the $\operatorname{P}(x_i|\textbf{C}_k)$ can be estimated based on the average probability of $x_i$ in the category $k$:
\begin{equation}
\operatorname{P}(x_i|\textbf{C}_k) = \frac{\sum_{n=1}^{n_k} x_{ik}}{n_k} = \frac{a_{ik}}{n_k}
\end{equation}
In addition, a Laplace smoothing is used to avoid the zero probability problem. Upon each teach/correct action, the prior probabilities of all categories as well as the probabilities of ${x}_i$ in the category $k$, $\operatorname{P}({x}_i|\textbf{C}_k)$, are updated incrementally. It is worth to mention that Bayesian approaches are computational efficient since the parameter of the model can be updated upon a new data point is added. Moreover, they are memory efficient as new training instances are used to update category models and then forgotten immediately. We have considered a probabilistic classifier to map the representation of a given object, $\textbf{x}^*$, to a label, $f_t(\textbf{x}^*) = \hat{y_i}$, through the maximum likelihood, ${\operatorname{argmax}}_k~\operatorname{P}(\textbf{C}_k|\textbf{x}^*) = \log \operatorname{P}(\textbf{C}_k) + \sum_i^d x^*_i ~ \log \operatorname{P}(x_i|\textbf{C}_k)$. 

\section{Experimental Results}
\label{exprimental_result}

\begin{figure}[!b]
    \centering
    \includegraphics[width=\linewidth, trim = 0cm 0cm 0cm 0cm clip=true]{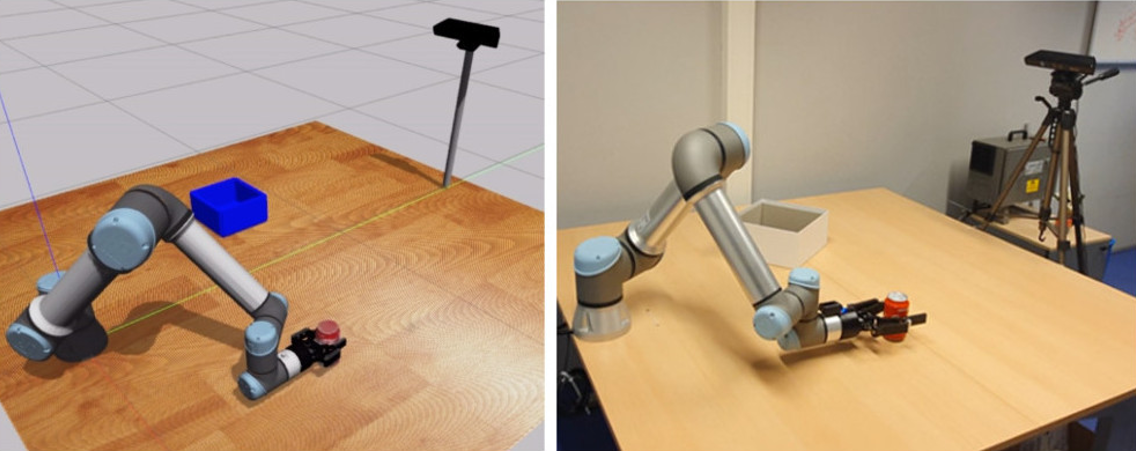}\vspace{0.3mm}\\
    \includegraphics[width=\linewidth]{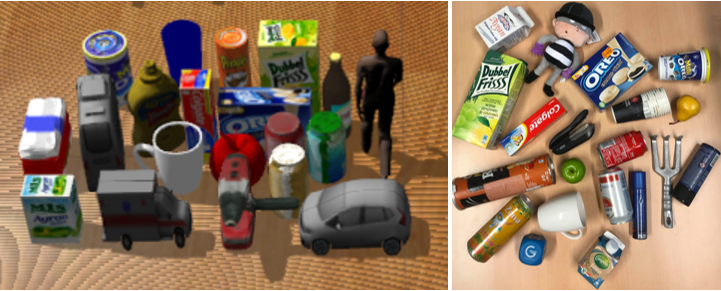}
    \caption{\small Our experimental setups: (\textit{top-left}) simulation environment in Gazebo; (\textit{top-right}) real-robot setup. The pose of the robot and the camera in the simulation are consistent with the real-robot setup. (\textit{bottom-left}) Objects used in simulation experiments; (\textit{bottom-right}) Objects used in real-robot experiments.}
    \label{grasp_setup}
    \vspace{-1mm}
\end{figure}

We evaluated our approach in both simulation and real-robot settings. Our setup consists of a Kinect camera, a Universal Robot (UR5e) with a two fingered gripper (Robotiq 2F-140), and a user interface. It should be noted that the pose of the robot and the camera in the simulation are similar to the real-robot setup (see Fig.~\ref{grasp_setup} \textit{top-row}).
We used a set of $20$ simulated objects, imported from the YCB dataset~\cite{calli2017yale} and Gazebo repository, and another set of $20$ real daily-life objects with different shapes, materials, sizes, texture, and weight (see Fig.~\ref{grasp_setup} \textit{lower-row}). More specifically, the selected objects have cubic, cylindrical, spheres, and special shapes and made out of cartons, iron, fabric, and plastic. All the objects used in real experiments were ``\textit{novel}'' and were not involved in the training procedure. 

We used the same code and network in both real and simulation experiments. Note that all tests were performed with a PC running Ubuntu $18.04$ with a $3.20$ GHz Intel Xeon(R) $i7$, and a Quadro P$5000$ NVIDIA. 

\subsection{Multi-view Grasp Dataset Generation}

\begin{figure}[!t]
    \centering
    \includegraphics[width=\linewidth]{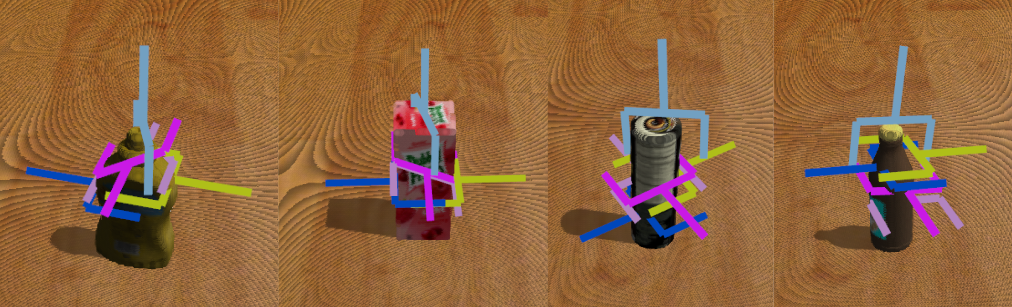}\vspace{0.5mm}\\
    \includegraphics[width=\linewidth]{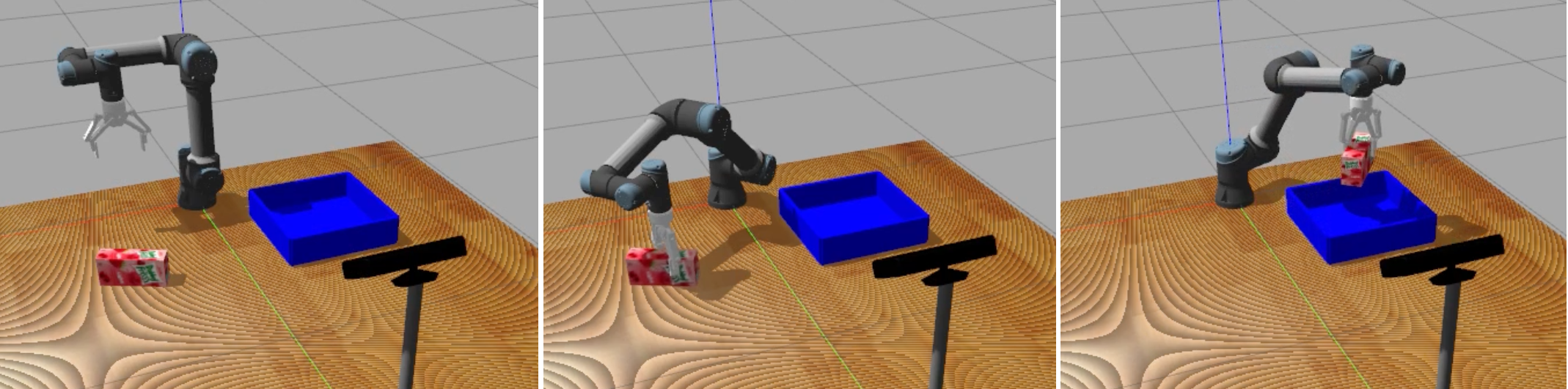}
    \caption{\small Multi-view grasp dataset generation: (\textit{top-row}) Examples of generated grasp synthesis for various objects; (\textit{lower-row}) Sequence of snapshots taken from one of the performed simulation experiments: ({left}) we randomly place an object (e.g., juice box) in the workspace of the robot; ({center})  The robot iteratively selects one of the extracted views of the object to approach and grasp the object; ({right}) the selected grasp synthesis is considered as a positive sample if the robot could pick and place the object into the basket.  We record all positive samples and discard those configurations that lead to a collision with the object or the table (negative samples).
    }
    \label{fig:dataset_setup}
\end{figure}

In order to generate a synthetic dataset, we randomly spawn an object in the workspace of the robot as shown in Fig.~\ref{fig:dataset_setup}. The robot then detects the object and extracts the multiple views of the object. In order to obtain a ground truth grasp configuration, we randomly sample grasp configurations for each of the extracted view of the object. 
We then convert each grasp configuration to 3D space and optimize the selected grasp configuration using simulated annealing~\cite{kirkpatrick1983optimization} by iteratively updating the orientation and width of the gripper. We compute a fitness value for the optimization process based on three main factors: (\textit{i}) the proportion of object's points that are between the gripper's fingers relative to all object's points (coverage criteria); (\textit{ii}) how stable the point is, which is measured based on how well the normals of the fingers overlap with the normals of the selected points between the two fingers; and (\textit{iii}) we also considered the distance of the selected grasp point to the center of projected view. Examples of generated grasp synthesis for different objects are depicted in Fig.~\ref{fig:dataset_setup} (\textit{top-row}). 

Furthermore, to make sure that the obtained grasp configuration is stable enough during manipulation phase, we instruct the robot to place the object into the blue basket (see Fig.~\ref{fig:dataset_setup}  \textit{lower-row}). To extend the size of dataset and cover various objects with different shape and size, we formed packed and pile of objects scenes using four to six objects and generate grasp configurations for those scenes in addition to generating grasp synthesis for isolated object scenario. Using the described procedure, we generate a grasp dataset of approximately one million positive grasp configuration and discard those configurations that lead to a collision with the object or the table (negative samples). 


\subsection{Ablation Study}
\label{network_analysis}

We trained several networks with the proposed architecture but different parameters including filter size, dropout rate, number of units in fully connected layers, loss functions, optimizer, and various learning rates, and batch size for $50$ epochs each. We used our synthetic grasp dataset to train the model. It should be noted that we augmented the data by zooming, random cropping, and rotating functions to generate approximately $5M$ grasp configurations. We trained the model on $80\%$ of data, and we kept $20\%$ for validation. We reported the obtained results based on the Intersection over Union (IoU) metric. A grasp pose is considered as a valid grasp if the intersection of the predicted grasp rectangle and the ground truth rectangle is more than $25\%$, and the orientation difference between predicted and ground truth grasp rectangles is less than $30$ degrees. We used Adam optimizer with a learning rate of $0.001$, Huber loss function ($\delta = 1.0$), and the batch size was set to $16$. The final architecture is depicted in Fig.~\ref{network_architecture}.

Furthermore, to study the effect of reconstruction loss on performance of object grasping, we trained the network wit and without reconstruction loss. We observed that the network without reconstruction loss achieved slightly better performance ($89.51\%$ vs. $89.24\%$). It is expected as by adding the reconstruction loss, we force the representation to include information that might be redundant for the grasping task. To encode the geometrical feature of the object for downstream recognition task, we used the output of the encoder part of the network, and discuss the effect of reconstruction loss on object recognition in the next section.

\subsection{Evaluations of Object Recognition}
Two rounds of experiments were performed to evaluate the proposed approach in offline and open-ended scenarios.

\subsubsection{Offline evaluation}

In this round of evaluation, a 10-fold cross-validation protocol (\texttt{train-then-test}) is used to assess the performance of the proposed approach. We used the Restaurant Object Dataset~\cite{kasaei2015interactive}, which contains 306 objects' views organizing in 10 categories with significant intra-class variations. Therefore, it is suitable for performing extensive sets of experiments. 
Our approach has several parameters that must be optimized to provide a good trade-off among recognition performance, memory usage, and computation time. The parameters are including: $\phi \in \{30^\circ, 45^\circ, 60^\circ\}$, $\alpha \in \{4, 8, \dots, 20\}, \beta \in \{ 3, 4, \dots, 7\}$, \texttt{view\_pooling} $\in$ \{max, avg, appending\}. The best results in terms of instance accuracy, class accuracy, and average computation time were found by running each possible permutation of the available parameters for \textit{Orthographic}, \textit{Orbit}, and \textit{Sphere} setups. To measure the performance of object recognition we used both instance accuracy ($acc_{micro}=\frac{\#\operatorname{true~predictions}}{\#\operatorname{predictions}})$ and average class accuracy ($acc_{macro} = \frac{1}{K} \sum_{i=1}^K{acc_i}$). Note that we report average class accuracy to address class imbalance, since instance accuracy is sensitive to class imbalance. In addition, we evaluated the effect of various input modalities on object recognition, including: depth-only, RGB-only (embedding of ViT is used as object representation), and RGB-D (the concatenation of embedding layers of Grasp network and ViT is considered as object representation). Refer to Table~\ref{table:result_or} for a summary of the best results for each camera setup. 

\begin{table*}[!b]
	\begin{center}
		\caption {\small Summary of offline evaluations for different input modality and various camera setups. The best of each modality is highlighted in bold, and the second-best is denoted by italicized text.}
		\vspace{-1mm}
\resizebox{\linewidth}{!}{
\begin{tabular}{ |c|c|c|c|c|c|c|c|c|c| }
    \hline
        \multirow{2}{*}{\textbf{Camera setup}} & \multicolumn{3}{c|} {Depth-only} & \multicolumn{3}{c|} {RGB-only} & \multicolumn{3}{c|} {RGB-D} \\\cline{2-10} 
     &  {\textbf{Orthographic}} &  {\textbf{Orbit}}  &  {\textbf{Sphere}} &  {\textbf{Orthographic}} &  {\textbf{Orbit}}  &  {\textbf{Sphere}}  & {\textbf{Orthographic}} &  {\textbf{Orbit}}  &  {\textbf{Sphere}}  \\\hline
     \textbf{\#Views} &  3  &  20  &  28 &  3  &  12  &  20 &  3  &  12  &  24 \\\hline
     \textbf{Pooling} &  Avg &   Max &  Max &  Max &   Max &  Max& Max &  Max &  Max \\\hline
     \textbf{Instance accuracy} &  0.9511 &  \textbf{0.9674} &  \textit{0.9642} & 0.9609 & \textit{0.9642} & \textbf{0.9674} & \textit{0.9674} & 0.9511 & \textbf{0.9772}\\\hline
     \textbf{Avg. class accuracy} &  0.9366 &  \textbf{0.9588} & \textit{0.9406} & 0.9400 & \textit{0.9424} & \textbf{0.9531} & \textit{0.9588} & 0.9324 & \textbf{0.9611} \\\hline 
     \textbf{Avg. computation time} (s) &  \textbf{{0.0167}} &  0.1102 &  0.1540 &  \textbf{{0.0214}} &  0.1574 &  0.2623 & \textbf{{0.0381}} &  0.2235 &  0.4470  \\\hline
\end{tabular}}
	\label{table:result_or}
	\end{center}
\end{table*}

\begin{figure}[!b]
    \centering
    \includegraphics[width=0.95\linewidth]{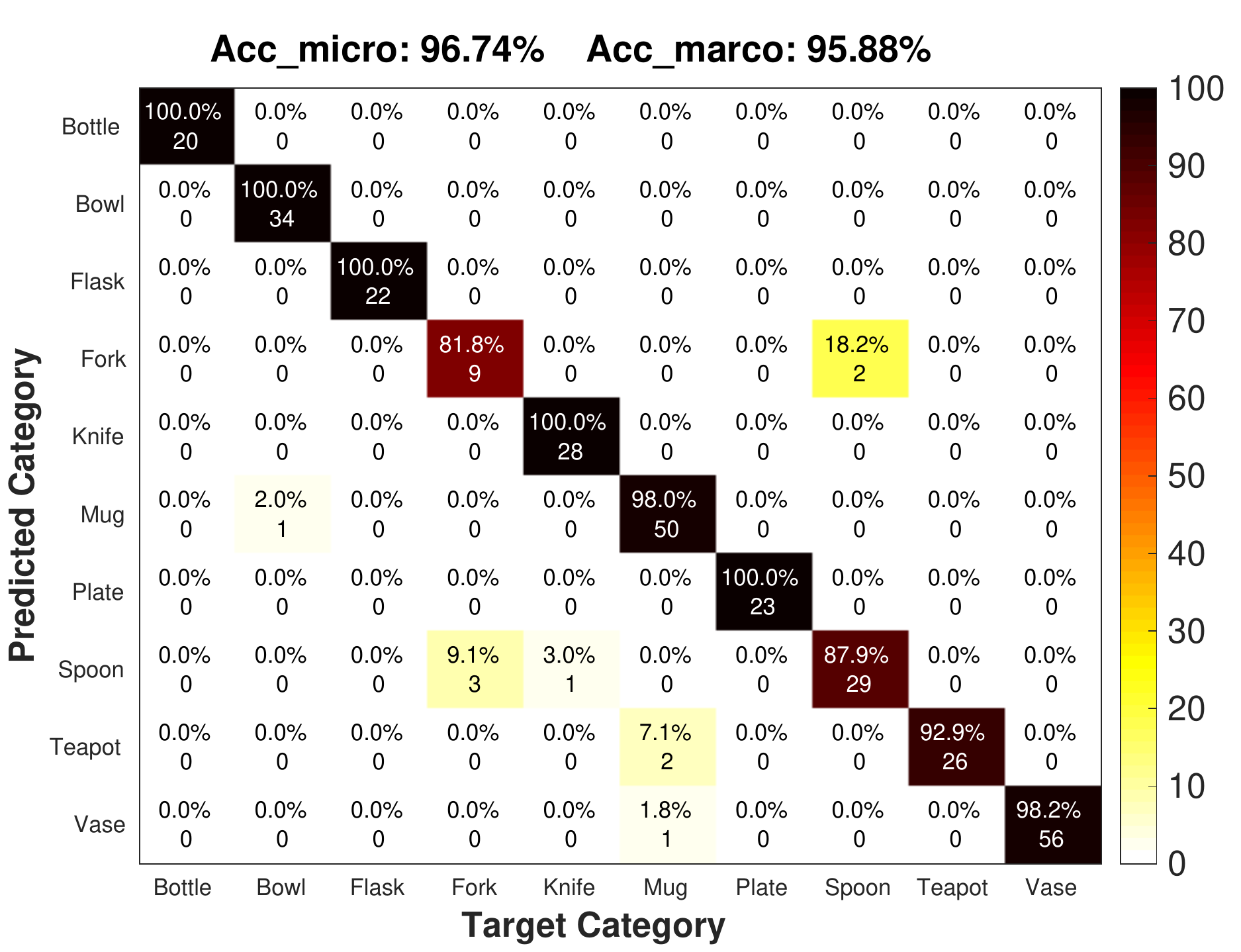}
  \caption{\small Confusing matrix for the best depth-only orbit setup ($\phi = 60^{\circ}$): based on this matrix, it is visible that the most of the misclassification happened among fine-grained categories (i.e., Fork, Knife, Spoon).}
  \label{cfx}
\end{figure}

\noindent\textbf{Depth-only:} 
In this round of experiments, we fed the depth views of the object into the encoder part of the Grasp network, and considered the output of the embedding layer as the feature vector. Results are reported in the \textit{depth-only} part of the Table~\ref{table:result_or}.
By comparing all the results, we observed that the best results obtained by the Orbit setup with $20$ views, and max pooling. We also visualized the confusion matrix for this setup in Fig.~\ref{cfx}. It is evident that most of the misclassification happened between objects that were extremely similar to one another (e.g., fork vs. spoon). This issue can be addressed through a fine-grained object categorization~\cite{keunecke2020combining,ji2020attention}. By comparing all experiments, it is visible that Orbit ($\phi = 60^{\circ}$) and Sphere ($\alpha=7,\beta=4$) setups achieved slightly better instance and average class accuracies than Orthographic setup ($\sim2\%$). 

\begin{figure}[!t]
    \centering
    \includegraphics[width=\linewidth, trim={0  0mm 0 0cm}, clip=true ]{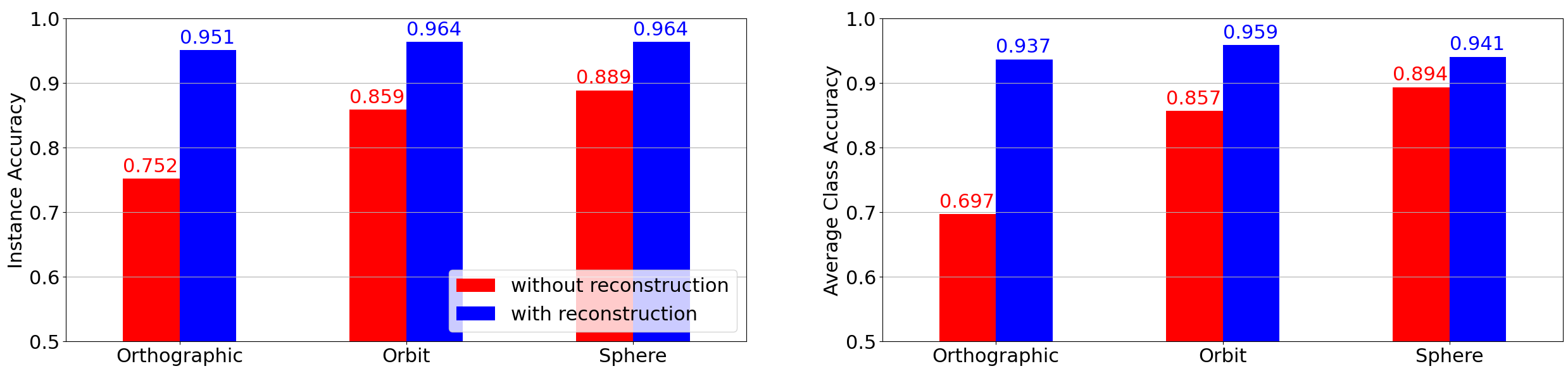}
    \caption{The effect of reconstruction loss on object recognition performance on Restaurant and Washington datasets. For a comprehensive evaluation, we have evaluated all virtual camera setups (orthographic, orbit, and sphere).}
    \label{fig:recognition_accs}
    \vspace{-4mm}
\end{figure}

Another set of experiments was conducted with the network without reconstruction loss to check the effect of reconstruction loss on object recognition accuracy. Results are depicted in Fig.~\ref{fig:recognition_accs}. By comparing the obtained results, it is visible that the reconstruction loss contributes significantly to learning descriptive representation. In particular, in all view setups, our network with reconstruction loss produced richer representations that led to better performance. In the case of Orthographic setup, both instance accuracy and average class accuracy are significantly improved (approximately $20\%$ and $24\%$) by using reconstruction loss. Similarly, in the case of Orbit and Sphere setups, the network with reconstruction loss outperformed the network without reconstruction loss concerning both instance accuracy and average class accuracy. In addition, we observed that the performance of the network without reconstruction loss increases as the number of views increases. This can be due to the fact that various views may activate different filters, and as a consequence, the obtained representation involves diverse features of the object which might improve the recognition performance.   

\begin{figure*}[!t]
    \centering
    \includegraphics[width=\linewidth, trim= 5.35cm 0.1cm 4.cm 0cm,clip=true]{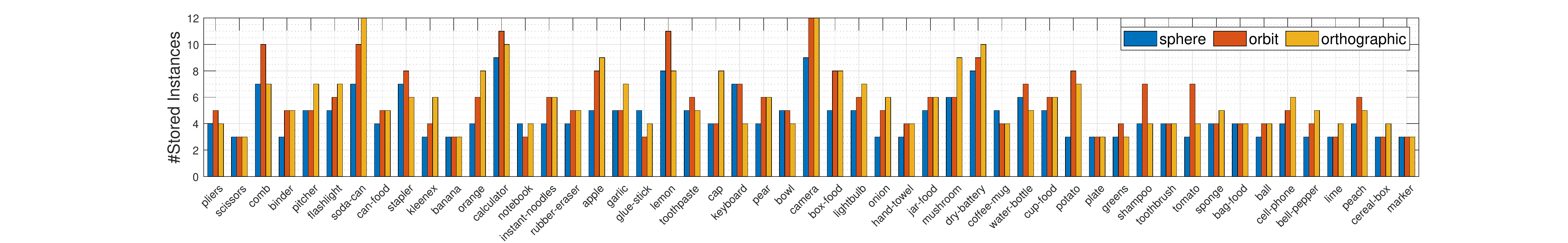}\\
    \begin{tabular}{cc}
        \hspace{-2mm}\includegraphics[width=0.5\linewidth, trim= 1cm 0.cm 1cm 0cm,clip=true]{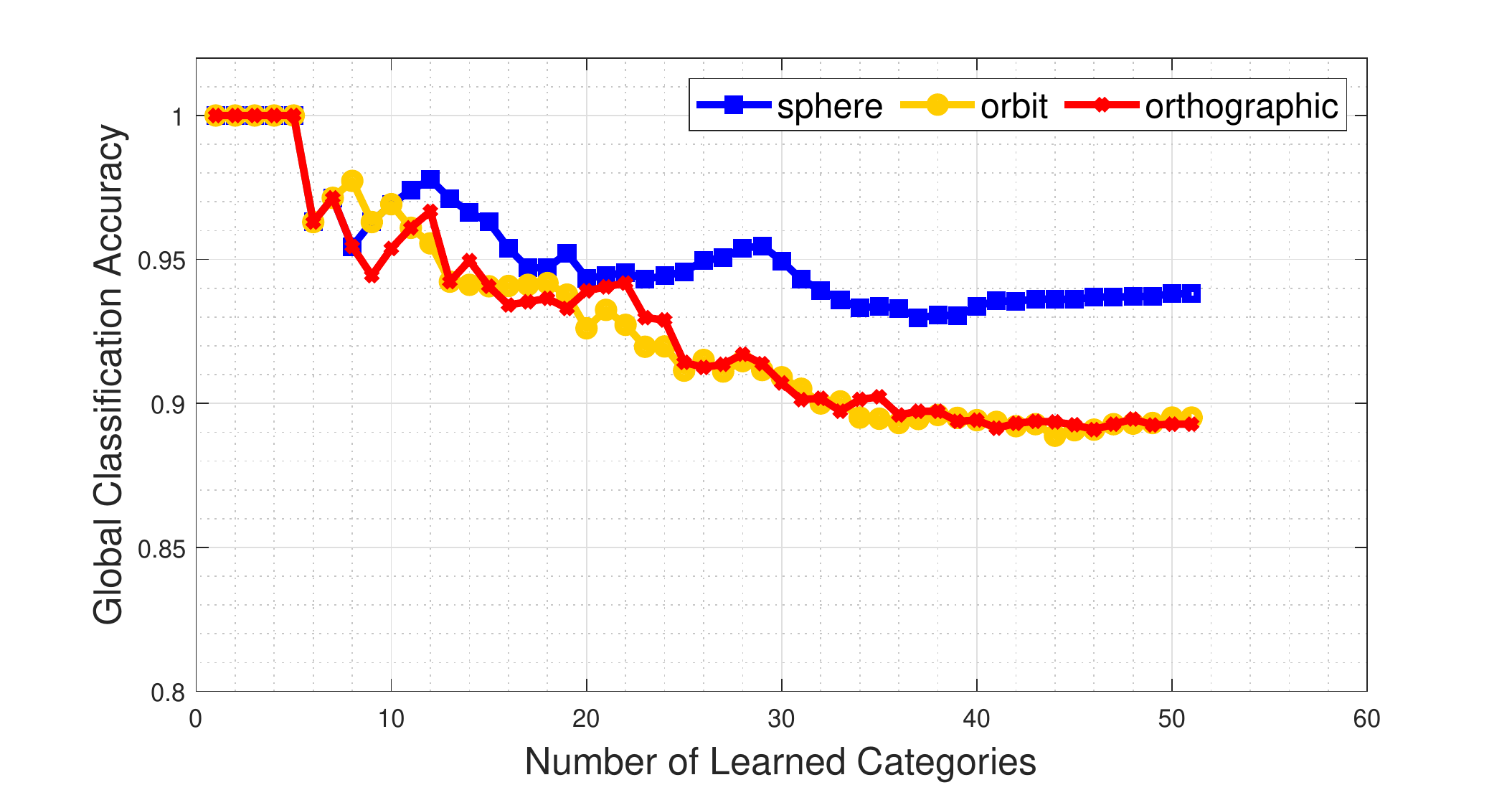}&
        \hspace{-3mm}\includegraphics[width=0.5\linewidth, trim= 1cm 0.cm 1cm 0cm, clip=true]{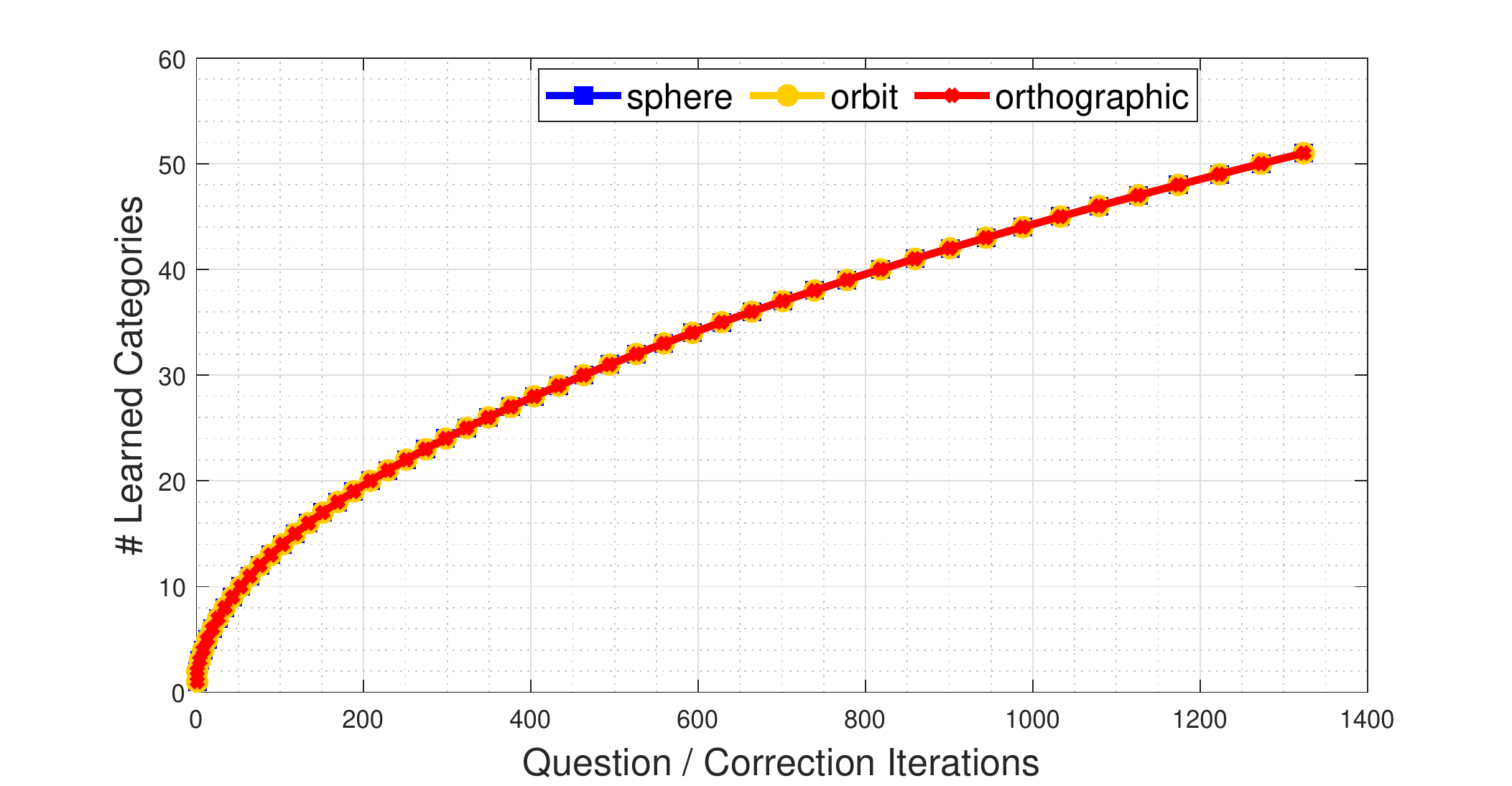}
    \end{tabular}
  \vspace{-3mm}
  \caption{System performance during the first simulated user experiment: (\textit{top}) This graph shows the number of instances used to train the object categories' model in orthographic, orbit, and sphere setups; (\textit{lower-left}) Global classification accuracy as a function of number of learned category; (\textit{lower-right}) Number of learned categories as a function of question/correct iterations.}
  \vspace{-2mm}
  \label{op_exp}
\end{figure*}

\noindent\textbf{RGB-only:}
In this round of experiments, 
we fed the RGB views of the object into the ViT, and considered the output of the global embedding layer as the representation of the object. 
Results are reported in the \textit{RGB-only} part of the Table~\ref{table:result_or}. Experimental results showed that the best recognition accuracies achieved by Sphere camera-setup using $20$ views distributed around object as $\alpha =5$, $\beta=4$. The second best result was achieved by Orbit setup by considering $12$ views of the object. In particular, orbit setup achieved $0.9642$ instance accuracy and $0.9424$ average class accuracy. Orthographic setup showed slightly worse instance and average class accuracies, $0.9609$ and $0.94$, respectively.

\noindent\textbf{RGB-D:}
To encode RGB-D views, we fed the best depth view into the Grasp network, and passed the RGB view to the ViT network. The obtained representations are then concatenated. To form a global representation for the object, all the views of the object are fused using a pooling function.  The right part of Table~\ref{table:result_or}  summarizes the results for various camera setups and RGB-D modality. By comparing all results it is clear that the Sphere setup achieved the best recognition accuracies using $24$ views. Interestingly, the orthographic setup achieved the second best recognition accuracies and the best computation time. In this round of experiments, orbit setup with 12 views obtained the third place. 

In the case of average computation time,
the orthographic setup outperformed orbit and sphere configurations with a large margin (see Table~II) regardless of input modality. This result shows that the orthographic setup can be used in closed-loop control ($\sim>25$Hz feedback) while orbit ($\sim>5$Hz feedback) and sphere ($\sim>2$Hz feedback) are computationally expensive for real-time applications. Therefore, we used the orthographic camera setup with RGB-D input modality for the real-robot experiments. 

\subsubsection{Open-ended evaluation}
We adopted an open-ended evaluation protocol that follows \texttt{test-then-train} scheme~\cite{chauhan2011using}\cite{kasaei2018coping}, to emulate the learning behaviour of a robot over long periods of time. In particular, it would be expected that the robot could be taught new categories that are present in its surroundings. It would be corrected on misclassifications it makes by a human user. Such experiments might take a long time with a human user. Therefore, we developed a \texttt{\small simulated user} to conduct systematic, consistent, and reproducible experiments. The simulated user can interact with the robot using \texttt{\small teach}, \texttt{\small ask}, and \texttt{\small correct} actions. We connect the simulated user to the largest publicly available 3D partial view object dataset~\cite{lai2011large} that contains $51$ object categories with $250,000$ views of $300$ objects.

In this round of experiments, the robot will start with no previous knowledge. The user teaches a category using three randomly selected views. After that, the user repeatedly picks unseen object views of the currently known categories and tests the robot to see if it has learned the category. This is done by asking the robot to identify new testing examples of all previously learned categories. When the agent makes a classification mistake, the user will provide feedback with the correct category label. This causes the robot to adjust its category model using the mistaken instance and also the prior probabilities of all categories are updated. The user estimates the recognition accuracy of the robot using a sliding window over the last $3n$ iterations, where $n$ is the number of categories. 
If the classification accuracy exceeds a threshold, $\tau = 0.75$, a new category is introduced. If the robot can not reach the classification threshold after 100 iterations since the last category was taught , the user realized that the robot is not able to learn more categories and terminates the experiment (breakpoint).  It is also possible that the robot learns all categories before reaching the breakpoint, and hence, the experiment is halted (reported as ``\textit{lack of data}'' condition)~\cite{chauhan2011using}\cite{kasaei2018coping}.

\noindent\textbf{Evaluation metrics}: Since the order of introducing the categories may matter, we run ten experiments for each approach and evaluate all approaches using five metrics as introduced in~\cite{chauhan2011using}\cite{kasaei2018coping}: an average number of learned categories (ALC), which shows \textit{how much the system is capable of learning}; the number of question/correction iterations (\#QCI) needed to learn those categories, and the average number of stored instances per category (AIC), \textit{shows the amount of time and memory needed for learning}; Global Classification Accuracy (GCA), representing the accuracy of agent computed based all predictions, and the Average Protocol Accuracy (APA), which represents the average accuracy of the agent over all sliding windows of the protocol. 

\begin{table}[!t]
	\begin{center}	
		\caption {Open-ended evaluations. The arrow demonstrates if better results are
higher or lower for each metric.}
            \resizebox{0.9\linewidth}{!}{
			\begin{tabular}{ |c|c|c|c|c|c| }
				\hline
				\textbf{Approaches} &
    		\#\textbf{QCI}$\mathbf{\downarrow}$ & \textbf{ALC}$\mathbf{\uparrow}$ & \textbf{AIC}$\mathbf{\downarrow}$ & \textbf{GCA}$\mathbf{\uparrow}$ & \textbf{APA}$\mathbf{\uparrow}$
				\\
				\hline \hline
				BoW~\cite{kasaei2018towards} & 724.30 &18.40 & 17.24 & 0.74 & 0.78\\
				\hline 
				Open-Ended LDA~\cite{hoffman2010online} & \textbf{572.10} &12.50 & 12.43& 0.73& 0.79\\
				\hline
				Local-LDA~\cite{kasaei2016hierarchical} & 872.10 &32.30& 11.58& 0.77 &0.81\\
				\hline
				GOOD~\cite{kasaei2018perceiving}& 1869.2 & 34.40 & 19.70 & 0.70 & 0.78 \\
				
				\hline \hline 
				ours-Orthographic & 1334.20 &	\textbf{51.00}	&	5.78	&	0.89 &	0.91\\
				\hline 
				ours-Orbit & 1329.40	&	\textbf{51.00} 	&	5.76	&	0.89	&	0.91\\
				\hline
				ours-Sphere & {1325.20}	&	\textbf{51.00}	&	\textbf{4.61}	&	\textbf{0.94}	&	\textbf{0.95}
				
				\\
				\hline
		\end{tabular}}
		\label{table:open_ended_evaluations}
	\end{center}
\end{table}

\noindent
\textbf{Results}: We compared our approach with four state-of-the-art methods. The obtained results are summarized in Table~\ref{table:open_ended_evaluations}. We also plot the performance of the proposed multi-view approaches in the first open-ended experiment in Fig.~\ref{op_exp}. By comparing all approaches, it is visible that \textit{sphere} camera setup outperformed orthographic and orbit configurations by a large margin in all evaluation metrics. The same results achieved when comparing our approach with the selected state-of-the-art approaches. In particular, the agent with multi view setup learned all existing categories in all experiments (ALC metric), and the stopping condition was ``\textit{lack of data}''. This result shows the potential for learning many more categories. The robot with \textit{orthographic} and \textit{orbit} camera setups achieved similar scalability by learning all $51$ categories. The other selected approaches, on average, learned less than $35$ categories and their performance drops aggressively when the number of categories increases. It is also clear that the robot with \textit{sphere} setup, on average, stored fewer instances per category, i.e., it required less than five instances per category while the other approaches, on average, need at least $5.76$ instances per category (AIC). It should be noted that \#QCI, GCA, and APA metrics should be seen in light of the number of learned categories. For instance, Open-Ended LDA achieved the best \#QCI, which is expected since it learned much fewer categories than our approaches (i.e., $12.50$). Hence, it can be concluded that our approach with multi view setup could learn all categories and outperformed all the selected approaches by a large margin. 

\subsection{Grasp Evaluations}
\label{grasp_results}
\begin{figure*}[!t]
    \centering
    \includegraphics[width=\linewidth]{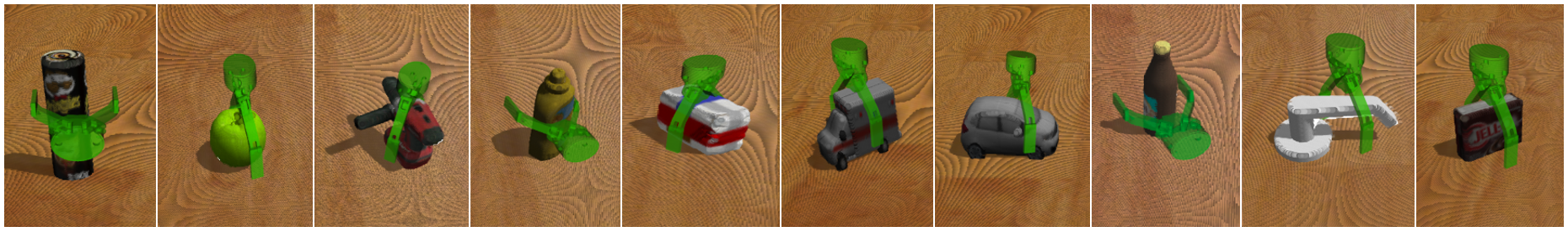}
    \vspace{-4mm}
  \caption{\small Visualizing the best grasp configuration for 10 household objects. The green grippers show the predicted grasp configurations. }
  \label{clear_table_task_isolated}
\vspace{-1mm}
\end{figure*}
In this round of evaluation, we designed a pick and place scenario in the context of a \textit{clear\_table} task. At the beginning of each experiment, we set the robot to a pre-defined setting, and randomly place objects on the table. In these experiments, the robot needs to learn, recognize, and detect the pose of the \textit{basket} as the placing pose, as well as the label and pose of another object to be cleaned from the table. Towards this goal, a user teaches the robot about the objects using a graphical menu. Afterward, the robot infers a graspable pose of the target object, picks it up, and puts it in the \textit{basket} (see Fig.~\ref{clear_table_task}). We performed this scenario not only to see whether the object slips due to bad grasp or not, also to show the coupling between grasping and recognition. We assess the performance of our approach in three scenarios, including isolated cluttered, pile of objects, and dense cluttered scenarios by measuring success rate, i.e., $\frac{\#success}{\#attempts}$. In this round of experiments, a particular grasp is considered a success if the object is inside the basket at the end of the experiment. 

\subsubsection{Isolated cluttered scenario}
\begin{figure}[!t]
    \centering
    \includegraphics[width=\linewidth]{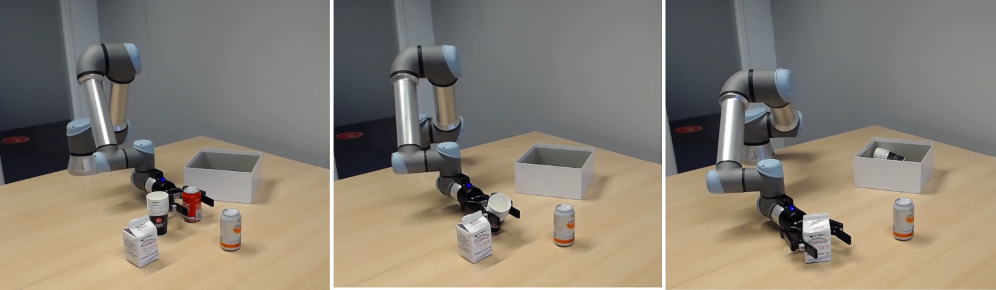}
    \vspace{-2mm}
    \caption{\small The sequence of snapshots taken from one of the real-robot experiments: we randomly place four objects in the workspace of the robot. The robot should pick and place objects into the basket one by one. In each iteration, the robot selects the nearest object to its base, chooses the best view of the object to infer grasp points for the object. To complete the task successfully, the robot executes pick and place actions to place the object into the basket. }
    \label{clear_table_task_real}
\end{figure}

Each simulated object was tested in isolation $50$ times, while each real-object was tested $5$ times. Note that, to speed up the real-robot experiments, we randomly placed four objects on the table to form a cluttered scenario first, and then instruct the robot to clean the object one by one (see Fig.~\ref{clear_table_task_real} and \ref{clear_table_task_isolated}). Therefore, the robot should recognize all objects precisely, and move them into the basket. 
In this round of experiments, we considered orthographic views to infer grasp configurations. These experiments can therefore be used as a stand-in for assessing the impact of view selection on grasping. We compared our approach against five baselines, including: Grasp Pose Detection (GPD)~\cite{gualtieri2016high} (an analytical approach), DexNet~\cite{mahler2017dex}, GG-CNN~\cite{morrison2018closing}, GR-ConvNet~\cite{kumra2020antipodal}, and Morrison et al.~\cite{morrison2020learning}. In our experiments, DexNet~\cite{mahler2017dex}, GG-CNN~\cite{morrison2018closing}, GR-ConvNet~\cite{kumra2020antipodal}, and Morrison et al.~\cite{morrison2020learning} have access to global projected top-down view of the full scene, while our approach uses extracted views of the object. The GPD method uses the partial point cloud of the object as input. Results are summarized in Table~\ref{table:isolated_exps}.

\begin{table}[!b]
\centering
    \vspace{-3mm}
    \caption{Evaluation of object grasping methods.}
    \label{table:isolated_exps}
    \resizebox{0.9\linewidth}{!}{
    \begin{tabular}{|c|c|c|}
    \hline
    \textbf{Method} & \textbf{Type} & \textbf{Success rate ($\%$)}\\
    \hline\hline
    GPD & sim  & 78.7 (787/1000) \\\hline
    GG-CNN & sim  & 72.6 (726/1000) \\\hline
    Morrison et al. & sim & 77.1 (771/1000) \\\hline
    DexNet & sim & 79.4 (794/1000) \\\hline
    GR-ConvNet & sim & 81.4 (814/1000)\\\hline
    Ours~(top-down) & sim  & 80.1 (801/1000) \\\hline
    Ours~(random) & sim  & 52.8 (528/1000) \\\hline
    Our & sim  & \textbf{91.8} (918/1000) \\\hline\hline
    GPD & real & 81.0 (81/100) \\\hline
    GG-CNN & real & 78.0 (78/100) \\\hline
    Morrison et al. & real & 77.0 (77/100)\\\hline
    DexNet & real & 81.0 (81/100) \\\hline
    GR-ConvNet & real & 81.0 (81/100) \\\hline
    Ours~(top-down) & real  & 82.0 (82/100) \\\hline
    Ours~(random) & real  & 61.0 (61/100) \\\hline
    Our & real & \textbf{92.0} (92/100) \\\hline
    \end{tabular}}
\end{table}

In the case of simulation isolated object experiments, the proposed approach achieved a grasp success rate of $91.8\%$ (i.e., $918$ success out of $1000$ trials), and for real objects, the success rate was $92\%$ ($92$ success out of $100$ attempts). By comparing all approaches, it is clear that the proposed approach significantly outperformed the selected approaches in both simulation and real-robot experiments (see Table~\ref{table:isolated_exps}).
More specifically, in the case of simulation experiments, the proposed approach worked $13\%, 19.2\%, 14.7\%, 12.4\%$, and $10.4\%$ better than GPD, GGCNN, Morrison et al., DexNet, and GR-ConvNet, respectively.

We found that the bulk of GGCNN, DexNet, and GR-CovNet errors were mainly due to estimating the center of the grasp point near the edge of an object. Therefore, as the gripper closes, the object may be pushed out. It should be emphasized that even very minor transformation errors might exacerbate these issues and cause the robot not to be able to grasp the target object. In contrast, since the proposed approach computes grasp configuration in object's reference frame, such failures did not happen to our approach.

We also observed that the success rate for GPD,
GGCNN, Morrision et al., in simulation experiments, was less than $80\%$, as they predict false positive grasp points and unsuccessfully attempted those grasp configurations. Such predictions often happened for small objects as it was not always possible to infer more than one grasp synthesis for them. Other failures were those brought on by insufficient friction, applying limited force to the object, running into other objects, and predicting unstable grasps synthesis.

Another interesting observation is that our approach with top-down view performed slightly better than GGCNN, GPD, Morrison et al., DexNet, in both simulation and real-robot experiments. We hypothesis that such differences come from this point that, since we placed several objects in the scene, GGCNN and GPD could infer a tiny space between two objects as a graspable area, leading to failures. In contrast, our approach considered local top-down view of the object. We also observed that, other failures were mainly happened in grasping \textit{SodaCan}, \textit{Colgate}, \textit{Fork}, and \textit{Toy}. Investigating the networks’ output reveals that the selected grasp point was in an unstable area where the surrounding area was too small, and therefore, the object slipped and fall during manipulation. Regarding our approach with random view selection, collision  with  the  table,  e.g.  grasping  a toppled soda-can from side, was the main reason of failure. Furthermore, we noticed that sometimes the robot was not able to find a kinematically feasible grasp point from the randomly selected view or top-down view.

We also observed that some of failures occurred when one of the fingers of the gripper was tangent to the surface of the target object, which led to pushing the object away. Other failures were mainly due to inaccurate object's bounding box, collision between the object and the basket (happened for large objects such as Pringles and JuiceBox). In the case of real-robot experiments, in addition to mentioned points, we found out some failures happened because of misclassification of the target and/or basket objects. In particular, as the robot placed more and more objects into the basket, the shape of the basket object was changed resulting in misclassification.


\subsubsection{Pile scenario}
\begin{figure*}[!t]
    \centering
     \includegraphics[width=\linewidth]{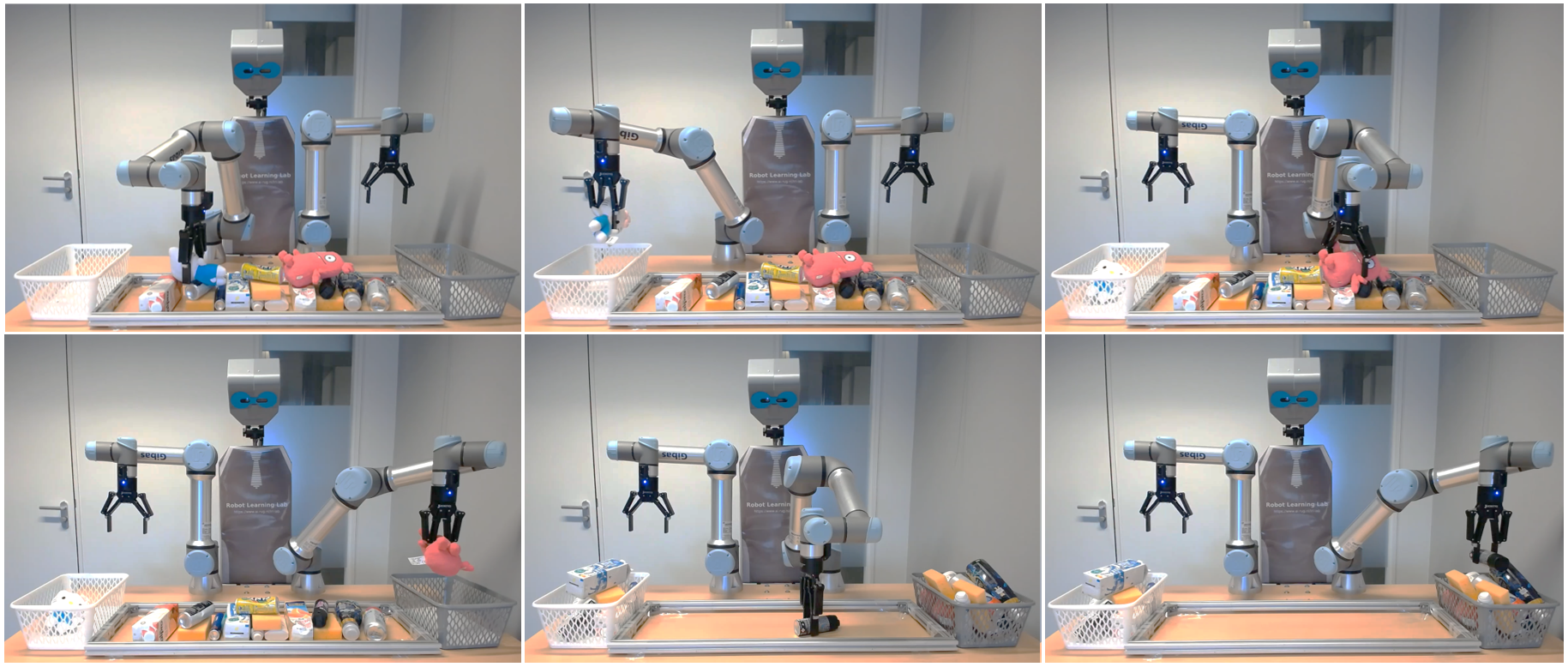}
  \caption{\small Object grasping in the highly cluttered scenario ($> 15$ objects): In this experiment, we make a pile of 15 objects in front of the robot and instruct the robot to perform a clear table task. The robot should then detect the grasp syntheses and execute the best grasp configuration. If the object is on the right side of the robot, the robot put the object into the right basket otherwise, the object is placed into the left basket by the robot. This procedure is repeated until all objects get removed from the table or five consecutive failures happen. }
  \label{highly_cluttred}
\end{figure*}
\begin{figure}[!b]
    \centering
     \includegraphics[width=\linewidth]{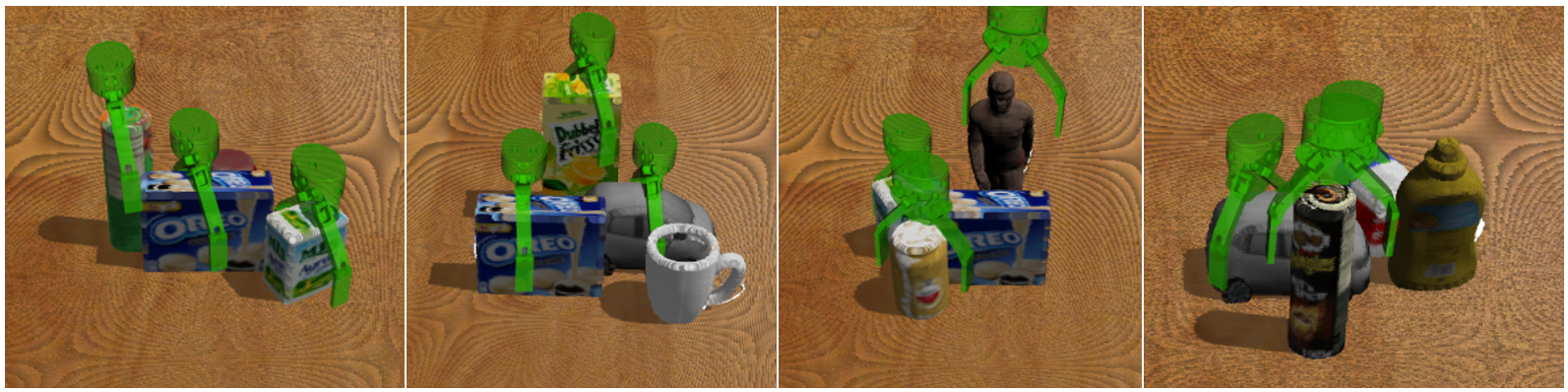}\\
    \includegraphics[width=\linewidth]{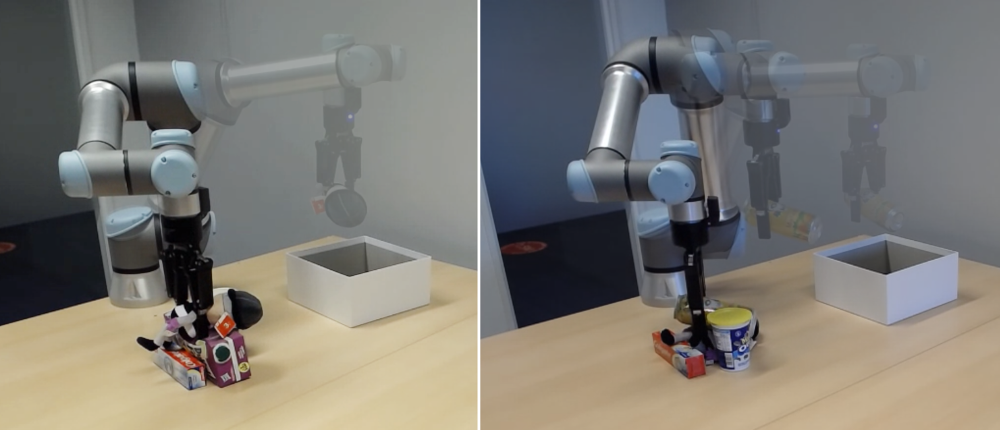}
  \caption{\small Qualitative results: (\textit{top-row}) visualizing the top-three grasp configurations on four simulated pile of objects; (\textit{lower-row}) The sequences of pick and place actions in two pile removal experiments. In order to complete the task successfully, the robot should pick and place all of the objects into the basket.}
  \label{clear_table_task_pile}
\end{figure}
  
We assess the performance of the proposed object grasping approach in pile scenarios. In this round of experiments, the robot knows in advance the pose of the basket, and needs to infer grasp points for the pile of five objects and put the objects into the basket one by one. An experiment is continued until either all objects get removed from the workspace, or three failures occurred consecutively. We performed $10$ real and $25$ simulated pile removal experiments. We visualized the top-three grasp predictions on four simulated pile of objects in Fig.~\ref{clear_table_task_pile} (top-row). The sequence of pick and place actions for two successful real experiments is shown in Fig.\ref{clear_table_task_pile} (lower-row). Regarding the simulation experiments, our approach could successfully removed $23$ out of $25$ pile of objects achieving $0.92$ pile removal. In the case of real-robot experiments, the robot could successfully complete the pile removal task in $9$ out of $10$ experiments, obtaining $0.90$ pile removal. We observed that the unreachable object was the underlying cause of the failure. In particular, when the robot was interacting with the pile of objects, one of the objects fell into a position that was not reachable by the robot. As a consequence, the experiment terminated after three consecutively failure attempts. Other reasons for failures were applying limited force to the object, colliding with another object, and predicting an unstable grasp. 

\subsubsection{Highly cluttered scenario}

In this round of experiments, we assess the performance of object grasping in highly cluttered scenario ($> 15$ objects) in the context of clear table task.
To accomplish this task successfully, the robot should detect the grasp syntheses and execute the best grasp configuration. If the object is on the right side of the robot, the robot uses its right arm and put the object into the right basket; otherwise, the object is placed into the left basket by the left arm of the robot. This procedure is repeated until all objects get removed from the table or five consecutive failures happen. A sequence snapshots showing the performance of the robot in such a scenario is depicted in Fig.~\ref{highly_cluttred}. It is worth to mention that, since we calculate the grasp synthesis in the object's reference frame, our approach is independent of the camera pose. 

In this round of evaluation, we performed $10$ experiments. The robot could successfully perform clear table task in $9$ experiments, and failed in one of the test. The underlying reason was that after removing some of the objects from the table, the spray object was situated next to the aluminum frame. The robot then tried to grasp the spray object, but it grasped both the object and the aluminum frame together, which led to failure. Finally, the experiment terminated after five consecutive failures happened. These experiments showed that the proposed object-agnostic grasp network was able to predict stable grasp configurations for novel objects including both isolated and piles of objects. A video of these experiments is available online at:  \href{https://youtu.be/n9SMpuEkOgk}{\cblue{\texttt{https://youtu.be/n9SMpuEkOgk}}}


\section{Conclusion}
\label{conclusion}
In this paper, we present a deep learning method to handle object recognition and grasping simultaneously. Our approach is especially suited for robots with limited resources. The proposed approach allows robots to incrementally learn new object categories and adapt to new environments by accumulating and conceptualizing experiments through interaction with non-expert human-users. We trained the proposed network in an end-to-end manner using a synthetic object dataset. As an input, the network receives a depth image and generates a deep representation encoding the geometrical feature of the object as well as pixel-wise grasp configuration as output. We fed the RGB views of the object into a ViT network to encode the textural feature of the object. We then concatenated the both RGB and depth feature vectors to form a global object representation. The obtained representation is finally used for open-ended object category learning and recognition through a meta-active learning technique. To validate the performance of our approach, we performed extensive sets of experiments in both simulation and real-robot. Experimental results showed that the overall object recognition and grasping performance of the proposed approach is significantly better than the best results obtained with the selected state-of-the-art approaches. Furthermore, the proposed approach allows robots to robustly interact with the environments in isolated object scenario, cluttered scenes, and pile of objects. In the continuation of this work, we would like to investigate the possibility of involving affordance mask as another output of the network for handling task-informed grasping (e.g., grasp the handle of a knife instead of its blade). 

{
\small
\bibliographystyle{ieeetr}
\bibliography{literature.bib}
}

\end{document}